\documentclass[conference]{IEEEtran}
\IEEEoverridecommandlockouts
\usepackage{cite}
\usepackage{amsmath,amssymb,amsfonts,amsthm}
\usepackage{multirow}
\usepackage[ruled,vlined,linesnumbered]{algorithm2e}
\SetKwInOut{KwIn}{Input}
\SetKwInOut{KwOut}{Output}
\usepackage{graphicx}
\usepackage{enumerate}
\usepackage{subfigure}
\usepackage{makecell}
\usepackage{enumitem}
\usepackage{ulem}
\usepackage{textcomp}
\usepackage{xcolor}
\usepackage{comment}
\usepackage[hidelinks]{hyperref}
\usepackage{float}
\usepackage[font=footnotesize,figurename=Figure]{caption}
\definecolor{mygreen}{RGB}{0,102,51}
\definecolor{myblue}{RGB}{0,51,102}
\definecolor{myorange}{RGB}{204,85,0}
\definecolor{myyellow}{RGB}{231,160,60}
\definecolor{myskyblue}{RGB}{79,173,235}
\definecolor{mypurple}{RGB}{68,8,125}

\newcommand{\defname}{Definition} 
\newtheorem{definition}{\defname}
\newcommand{\theoname}{Theorem} % 默认正文用的名字
\newtheorem{theorem}{\theoname}

\def\BibTeX{{\rm B\kern-.05em{\sc i\kern-.025em b}\kern-.08em
    T\kern-.1667em\lower.7ex\hbox{E}\kern-.125emX}}

\begin{document}
	
\renewcommand{\tablename}{R-Table}
\renewcommand{\figurename}{R-Figure}
\renewcommand{\theoname}{R-Theorem}
\renewcommand{\defname}{R-Definition}
\setcounter{section}{0}
\setcounter{figure}{0}
\setcounter{theorem}{0}
 \setcounter{definition}{0}
\setcounter{equation}{0}
\setcounter{table}{0}
\setcounter{page}{0} % Set the page number to start from 0
\renewcommand{\tablename}{Table}
\renewcommand{\figurename}{Figure}
\renewcommand{\theoname}{Theorem}
\renewcommand{\defname}{Definition}
%\title{AGRAG: A Graph-based Retrieval Augmented Generation Framework for Complex QA}

\title{AGRAG: Advanced Graph-based Retrieval-Augmented Generation for LLMs}

\author{\IEEEauthorblockN{Yubo Wang$^1$, Haoyang Li$^{2*}$\thanks{$^*$Corresponding Author}, Fei Teng$^1$, Lei Chen$^{1,3,4}$}
	$^1$The Hong Kong University of Science and Technology, Hong Kong SAR, China\\
	$^2$The Hong Kong Polytechnic University, Hong Kong SAR, China\\
	$^3$The Hong Kong University of Science and Technology (Guangzhou), China\\
	$^4$Guangzhou HKUST Fok Ying Tung Research Institute, China\\
	ywangnx@connect.ust.hk, haoyang-comp.li@polyu.edu.hk, fteng@connect.ust.hk, leichen@cse.ust.hk
}
\maketitle

\begin{abstract}
Graph-based retrieval-augmented generation (Graph-based RAG) has demonstrated significant potential in enhancing Large Language Models (LLMs) with structured knowledge. However, existing methods face three critical challenges: Inaccurate Graph Construction, caused by LLM hallucination; Poor Reasoning Ability, caused by failing to generate explicit reasons telling LLM why certain chunks were selected; and Inadequate Answering, which only partially answers the query due to the inadequate LLM reasoning, making their performance lag behind NaiveRAG on certain tasks. To address these issues, we propose AGRAG, an advanced graph-based retrieval-augmented generation framework. When constructing the graph, AGRAG substitutes the widely used LLM entity extraction method with a statistics-based method, avoiding hallucination and error propagation. During retrieval, AGRAG formulates the graph reasoning procedure as the Minimum Cost Maximum Influence (MCMI) subgraph generation problem, where we try to include more nodes with high influence score, but with less involving edge cost, to make the generated reasoning paths more comprehensive. We prove this problem to be NP-hard, and propose a greedy algorithm to solve it. The MCMI subgraph generated can serve as explicit reasoning paths to tell LLM why certain chunks were retrieved, thereby making the LLM better focus on the query-related part contents of the chunks, reducing the impact of noise, and improving AGRAG's reasoning ability. Furthermore, compared with the simple tree-structured reasoning paths, our MCMI subgraph can allow more complex graph structures, such as cycles, and improve the comprehensiveness of the generated reasoning paths.
The code and prompt of AGRAG are released at: \url{https://github.com/Wyb0627/AGRAG}.
	% }
%\footnote{\# haoyang: you should explicitly state how your method addresses existing challenges. Also,limited generalization without any solutions}
\end{abstract}


\section{Introduction}

Large Language Models (LLMs) \cite{sun2021ernie, jiang2023mistral, touvron2023llama, achiam2023gpt, yang2024qwen2} have demonstrated remarkable capabilities in natural language understanding and generation \cite{zhang2024finsql,chen2024data,lu2025adda,jo2025sparellm,balaka2025pneuma,zhou2025cracking,barbosa2025cost,luoma2025snails,omar2025dialogue}. 
However, their reliance on static, internalized knowledge often leads to factual hallucinations and limited adaptability to dynamic information \cite{gao2023retrieval,peng2024graph,wu2024retrieval,liu2024lost}, and retraining or finetuning the LLM can lead to high computational costs \cite{merth2024superposition,fan2024survey}.
Retrieval-Augmented Generation (RAG) addresses this by integrating external factual knowledge during inference, improving accuracy and enabling real-time updates. 
{% \color{blue}
Yet, traditional RAG methods' 
retrieval~\cite{lewis2020retrieval,gao2023retrieval} is based on simplistic mechanisms, such as whole passage's dense vector similarity, making them overlook the capture of more fine-grained semantic relationships between entities \cite{han2024retrieval}. }
% \footnote{\# haoyang: give what is the fine-grained, what is the meaning}
To overcome these limitations, researchers have proposed various Graph-based RAG models \cite{edge2024local,guo2024lightrag,listructrag,hu2024grag,gutierrez2024hipporag, liang2025kag,gutiérrez2025ragmemory,wang2025archrag,zhou2025depth}. 
% The framework of Graph-based RAG models, 
% GraphRAG~\cite{edge2024local}, 
% is shown in Figure \ref{fig:graphrag}. 
These models first utilize LLMs to extract entities and relations from the source documents to serve as graph nodes and edges and construct a graph, and then leverage this graph-structured knowledge to model complex relationships between entities to enable structured reasoning over inter-connected information, 
enabling the capture of more fine-grained entity-level relationships.
% enhancing both the relevance and coherence of retrieved evidences 
\cite{han2024retrieval}. 

Currently, the Graph-based RAG models can be classified into two categories, i.e., \textit{community-based models}, and \textit{walking-based models}. 
The community-based models \cite{edge2024local,hu2024grag,guo2024lightrag} 
first map the query to graph nodes considering their semantic similarity or keyword matching. 
Then, they will retrieve the mapped graph nodes' one or multi-hop neighborhood.
 % For example, GraphRAG \cite{edge2024local} creates graph communities, then maps the query to its corresponding communities and applies LLM to generate the community summary for retrieval. 
%  \footnote{\# why only graphrag, could you write one general approach and cite more ppaers}
On the other hand, with similar mapping approach from query to graph nodes, walking-based models \cite{gutierrez2024hipporag,fastgraphrag,listructrag,liang2025kag,gutiérrez2025ragmemory} apply walk-based graph algorithms, such as Personalized PageRank (PPR) \cite{haveliwala2002topic}, 
based on the mapped graph nodes, 
to score graph nodes and edges based on the graph's topological structure and query information, 
and then retrieve Top-\textit{k} text chunks based on high-scoring graph nodes, 
capturing the fine-grained entity level relationships. 
 %For example, when retrieval, HippoRAG2 \cite{gutiérrez2025ragmemory} first maps query extracted triples to graph nodes and edges based on semantic similarity, then applies PPR based on these mapped graph components, and finally retrieves Top-\textit{k} text chunks based on the PPR score of their linked nodes.

However, on complex QA tasks that require knowledge summarization or content generation \cite{xiang2025use}, 
the existing RAG models 
cannot answer these complex queries due to:
% still suffer from the following issues: 
\begin{enumerate}[leftmargin=*]
	{% \color{blue}
	\item \textbf{Inaccurate Graph Construction:} 
	Current Graph-based RAG models apply LLM for entity and relation extraction to construct graphs. However, the LLM can inevitably suffer from the hallucination issue \cite{gao2023retrieval,peng2024graph,wu2024retrieval,liu2024lost}, leading to inaccurate entity extraction, introducing noise into the constructed graph and propagate through the whole pipeline, affecting the model effectiveness.
	% Current Graph-based RAG models simply depends on LLMs' semantic ability for various procedure graph indexing, such as {\color{red}entity extraction} and
	% {\color{red}xxx}, which can lead to frequent LLM calls and cause significant computational overhead as well as high token cost. 
	
	\item \textbf{Poor Reasoning Ability:} 
	% {\color{red}
		% \footnote{\#haoyang: how about wander-based? Yubo: Fixed}
	The community-based models' retrieval only focuses on a narrow graph neighborhood and misses the graph's long-hop dependency, 
	which is crucial for tasks that require multi-hop reasoning. 
	In contrast, walking-based models do account for these long-hop dependencies during retrieval; however, they simply feed the retrieved text chunks directly into the LLM as contexts. 
	Consequently, both of their retrieval and reasoning processes remain opaque to the LLM, which receives no explicit structural reasoning chain to explain why those particular chunks were selected, making the LLM struggle to locate the query-related context within those chunks, leading to suboptimal query answering performance on certain tasks \cite{xiang2025use,han2025rag,zhou2025depth}.
	%  then solely depends on LLM for reasoning, 

	\item \textbf{Inadequate Answer:} 
	Although we can simply combine the existing RAG models with the current shortest path-based graph algorithm to provide explicit reasoning chains to aid LLM query answering, improve the model's reasoning ability. 
	However, on tasks that require summarization of current knowledge \cite{xiang2025use,han2025rag,zhou2025depth}, this simple approach can lead to suboptimal performance, as the shortest path on the graph can be overly simple, and can omit many query-related facts, leading to incomplete reasoning chains, which makes the model tend to generate inadequate query answer with low information coverage and faithfulness (See Table \ref{tab:abl_exp} and Section \ref{ssec:exp:ablation}).

	% According to recent researches \cite{xiang2025use,han2025rag,zhou2025depth}, 
	% on tasks that require summarization of current knowledge, the answer of current Graph-based RAG models tend to have lower coverage and faithfulness than simple NaiveRAG \cite{gao2023retrieval}, 
	% the performance of these Graph-based RAG models is even sub-optimal than the simple NaiveRAG \cite{gao2023retrieval} on tasks that require summarization of current knowledge, 
	% as their responses are not 
	% which implies their retrieval modules are not comprehensive enough.
}
\end{enumerate}

{
	% \color{blue}
To address these issues, we propose a novel 
\textbf{A}dvanced \textbf{G}raph-based \textbf{R}etrieval-\textbf{A}ugmented \textbf{G}eneration
%\textbf{G}raph-based \textbf{E}fficient \textbf{R}etrieval \textbf{A}ugmented \textbf{G}eneration 
framework, called AGRAG. 
We divide the AGRAG framework into three steps: 
% Furthermore, AGRAG also applies a statistic-based Named Entity Recognition (entity extraction) method for graph indexing to substitute HippoRAG2's LLM entity extraction approach, which consequently leads to less token and time cost while not affecting the model's effectiveness. We divide the AGRAG framework into three steps: 
In Step 1 Data Preparation, 
AGRAG first extracts entities with a TFIDF \cite{tfidf} based entity extraction method, avoiding LLM hallucination, and resulting in better graph construction accuracy.
Then, AGRAG applies LLM to detect the possible relation w.r.t. the extracted entity and the text chunk, formulating entity-relation-entity triples, namely knowledge triples.
Then these knowledge triples, as well as the text chunks involved, together formulate a Knowledge Graph (KG). 
% In this step, we design a light-weighted, statistics-based entity extraction method to substitute other baselines' LLM-based entity extraction approach, achieving better efficiency while not affecting the effectiveness.
% \footnote{\# haoyang: again, you have repeated it in last paragraph. And what is the method?}
In Step 2 Graph Retrieval, AGRAG maps the query to knowledge triples in KG based on their semantic similarity, then calculate the PPR score \cite{haveliwala2002topic} as node influence based on the KG structure and the mapped triples. In the meantime, AGRAG also calculates edge weights based on the semantic similarity between the query and the edge's corresponding triple. 
After that, AGRAG generates a Minimum Cost and Maximum Influence (MCMI) reasoning subgraph via a greedy algorithm, which is calculated based on the shortest spanning tree of the mapped triples. 
% is a extension of the Mehlhorn algorithm \cite{mehlhorn1988faster} for Steiner Tree calculation. 
It tries to maximize the node influence score, while minimize the edge cost in the expansion of each node, in order to avoid introducing noise. 
Compared to other shortest path-based algorithms, the MCMI subgraph allows for more complex structures, such as rings, which results in reasoning chains containing more query-related entities and relations, making them more comprehensive. This, in turn, helps the model perform better on complex tasks that require summarization, resulting in better reasoning ability and more comprehensive query answers.
% \footnote{\# very unclear, what is anchors the input query, could you directly state your method}
In Step 3 Hybrid Retrieval, 
% When generating the MCMI, we try to include nodes with high influence scores to improve the comprehensiveness while trying to minimize the total edge cost to avoid possible noise.
 %After retrieval, 
% the MCMI subgraph serves as explicit structural reasoning chains, along with its associated text chunks 
% maximizing query-relevant information for any given task while minimizing noise during the retrieval process without the need for any parameter finetuning. 
 %Compared to the simple steiner tree, the MCMI subgraph can contain more complex reasoning structures, such as cycles between key nodes in graph, make it more generalizable to complex scenarios and tasks. 
% We formally prove the problem of generating an MCMI subgraph is NP-hard, and AGRAG addresses it using a greedy algorithm.
AGRAG carries out hybrid retrieval based on the average of dense representation vector similarity score and sparse BM25 \cite{robertson2009probabilistic} keyword matching score between query and text chunks, providing extra contexts for query answering. 
At last, AGRAG will input the resulting subgraph from the Graph Retrieval step and chunks from hybrid retrieval to LLM for final query answering.
% After MCMI generation, the resulting subgraph, along with the corresponding text chunks it references and 
% the text chunks most semantically similar to the query, are then fed into the LLM for final query answering.
% \footnote{\# no any informative details for your method. It is too general, too redundant}
}

We summarize our novel contributions as follows:
\begin{itemize}[leftmargin=*]
	\item We present the AGRAG framework, which consists of three steps: Data Preparation, Graph Retrieval, and Hybrid Retrieval, aiming to address the Inaccurate Graph Construction, Poor Reasoning Ability, and Inadequate Answer issues of current Graph-based RAG models. 
	\item We design a statistic-based entity extraction method, avoiding the hallucination caused by LLM entity extraction, and improving the pipeline effectiveness.
	% to substitute for the costly LLM entity extraction procedure, and improve the model's efficiency while not affecting its effectiveness.
	\item We propose the MCMI subgraph generation problem, prove it to be NP-hard, and design a greedy algorithm, ensuring AGRAG's generalizability to complex tasks. 
	% \item To solve the MCMI subgraph generation problem, we design a greedy algorithm, generating a complex and clear reasoning subgraph through the mapped triples, ensuring AGRAG's generalizability to complex tasks. 
	\item According to our experiments, 
	AGRAG achieves better effectiveness and efficiency compared with current state-of-the-art RAG models on six tested tasks. 
	% can achieve at most 13.6\% effectiveness gain, 1.66x faster and 3.69x token saving than current Graph-based RAG models.
\end{itemize}
% \footnote{\# the introduction is too vague}

% However, the Graph-based RAG models still suffer

\begin{comment}
\begin{figure}[t]
	\centering
	\includegraphics[width=1\linewidth]{figures/DFSTC_task}
	\vspace{-10px}
	\caption{An example of the DFSTC task with two rounds. 
		In round 1, an example text  is classified as \textit{Machine Learning}.
		In round 2, an  example text is classified as \textit{Computer Vision}.
		% assigned by our edge weighting mechanism. 
	}
	\label{fig:dfstc_task}
	\vspace{-15px}
\end{figure}
\end{comment}

% \vspace{-15px}
\section{Related Works}\label{sec:pre_rel}
% {\color{blue}
	In this section, we first introduce the traditional RAG models in Section~\ref{sec:RAG_model} and then discuss the more advanced Graph-based RAG models in Section~\ref{sec:GRAG_model}. 
	The important notations used in this paper are listed in Table \ref{tab:notation}.
	%  \autoref{sec:notations}.
	
	% }

% Notation tables
% C_n^m new classes at round m
% C_b base classes at the begining

% \vspace{-10px}
\subsection{Traditional RAG Models} \label{sec:RAG_model}
Recently, Large Language Model (LLM)-based models have undergone rapid development 
\cite{fernandezEFKT23, amerYahiaBCLSXY23, zhangJLWC24, miaoJ024,snyder2024early,zhong2024logparser,yan2024efficient,chen2024large,ding2024enhancing}
and have been successfully adapted to various database related tasks, 
including 
data discovery \cite{arora2023language, dong2023deepjoin, kayali2024chorus}, 
entity or schema matching \cite{zhang2023schema, tu2023unicorn, fan2024cost, du2024situ}, and natural language to SQL conversion \cite{li2024dawn, ren2024purple, trummer2022bert, gu2023few}. 
% *%\footnote{Add KDD tasks and citations}
% including data discovery \cite{arora2023language, dong2023deepjoin, kayali2024chorus}, 
% entity or schema matching \cite{zhang2023schema, tu2023unicorn, fan2024cost, du2024situ}, and natural language to SQL conversion \cite{li2024dawn, ren2024purple, trummer2022bert, gu2023few}. 

Although LLMs are inherently capable of inference without fine-tuning \cite{sun2021ernie, jiang2023mistral, touvron2023llama, achiam2023gpt, yang2024qwen2}, the lack of fine-tuning with up-to-date and domain specific knowledge can lead LLMs to generate incorrect answers. 
As they lack task-specific knowledge, these incorrect answers are often referred to as hallucinations \cite{zhang2023siren}.
To mitigate hallucinations, researchers try to directly retrieve text documents outside of LLM based on the query as side information to help the LLM query answering 
% have provided LLMs with text formed side information for classification 
\cite{chen2023dense,sarthi2024raptor,hu2024prompt,wu2024coral,cai2024forag,oosterhuis2024reliable,yadav2024extreme,che2024hierarchical,zhou2025emorag, wang2024corag}, lead to the emergence of early RAG models, namely traditional RAG models. 
The earliest LLM-RAG model is the simple NaiveRAG \cite{gao2023retrieval}, when indexing, it splits the texts into chunks, and encodes them to semantic vector representations. When retrieval, the model retrieves several text chunks with top semantic similarity to the query. 

Based on NaiveRAG, researchers have proposed various walking-based models to further tailor the text chunks and improve the retrieval quality. 
For example, Propositionizer \cite{chen2023dense} applies a fine-tuned LLM to convert the text chunks into atomic expressions, namely propositions, to facilitate fine-grained information retrieval. 
Prompt compressor models \cite{pan-etal-2024-llmlingua,jiang-etal-2024-longllmlingua}, such as LongLLMLingua \cite{pan-etal-2024-llmlingua} apply LLM's generation perplexity to filter out unimportant tokens in the text chunks, try to maintain a more concise retrieval result for LLM input. 
However, the retrieval procedures of these models remain unstructured and struggle to capture the nuanced semantic and structured relationship within the text chunks, which impedes the effectiveness of LLMs \cite{gao2023retrieval,peng2024graph,wu2024retrieval,liu2024lost}.

\subsection{Graph-based RAG Models}\label{sec:GRAG_model}
To overcome the aforementioned limitations of traditional RAG models, researchers have begun to develop Graph-based RAG models \cite{edge2024local,guo2024lightrag,listructrag,hu2024grag,gutierrez2024hipporag, liang2025kag,gutiérrez2025ragmemory}. These models usually extract graph components with LLMs from text corpora to index them into graphs. When retrieval, these models will utilize the graph topology structure to capture complex relationships between entities within and across text chunks.

\subsubsection{\textbf{Tree-based Models}}
To address the limitation of flat retrieval in capturing long-range dependencies and thematic information, tree-based models organize text chunks into hierarchical structures. RAPTOR \cite{sarthi2024raptor} recursively clusters and summarizes text chunks to construct a tree from the bottom up, enabling the retrieval of information at varying levels of abstraction to answer high-level thematic queries. 
Building on this, SiReRAG \cite{zhang2025sirerag} constructs dual index trees to capture both semantic similarity and entity-based relatedness. It builds a similarity tree via recursive summarization and a relatedness tree that groups fine-grained propositions sharing the same entities, thereby indexing both similar and related information to support complex multihop reasoning.

However, these tree-based models can only generate tree-structured reasoning chains, which restricts their adaptivity to different tasks.

\subsubsection{\textbf{Community-based Models}}
Early Graph-based RAG models typically retrieve specific neighborhoods from the graph based on the input query. The pioneering work in this category is GraphRAG \cite{edge2024local}, 
% as illustrated in Figure \ref{fig:graphrag}, 
it aggregates nodes into communities and generates community summaries to encapsulate global information. 
To reduce the computational overhead of community detection, LightRAG \cite{guo2024lightrag} retrieves semantically similar entities along with their 1-hop graph neighborhoods and corresponding text passages, eliminating the need for pre-computed summaries.

Moving beyond fixed neighborhoods, G-Retriever \cite{he2024g} formulates subgraph retrieval as a Prize-Collecting Steiner Tree (PCST) optimization problem, identifying connected subgraphs that maximize relevance while minimizing costs. Recently, MIXRAG \cite{liu2025mixrag} introduces a Mixture-of-Experts (MoE) framework to handle diverse query intents, dynamically routing queries to specialized entity, relation, or subgraph experts and employing a query-aware graph encoder to filter noise.

Moreover, all these community-based models can be limited in capturing information beyond a small local neighborhood, thereby restricting their long-hop reasoning ability \cite{luo2024rog,zhou2025breaks}.

\begin{table}[t]
	% \color{blue}
	\centering
	\caption{Summary of important notations. 
		% We denote the vector representation of an item by its \textbf{bold} character.
		}
	\label{tab:notation}
	\renewcommand{\arraystretch}{1.05}
{
		\begin{tabular}{l|p{5cm}}
			\hline\hline
			
			\textbf{Notations}   & \textbf{Meanings}  \\ 
			\hline
			$t$ & The text chunk.\\ \hline
			
			$q$ & The input query. \\ \hline
			
			$\mathcal{T}$, $\mathcal{T}_c$ & Original texts and split text chunks. \\  \hline
			
			$l_t$, $l_o$ & Text chunk and token overlap length.   \\ \hline
			
			$s$, $c$ & The node influence score and edge cost.  \\ 		\hline
			$\mathcal{S}_\mathcal{V}$, $\mathcal{C}_\mathcal{E}$ & Set of node influence score and edge cost. \\ \hline
			
			$\tau$, $\phi$ & Entity extraction and synonym threshold.   \\ \hline
			$b$ & The maximum n-gram in extity extraction.   \\ \hline
			$k_a$, $k_r$ & Number of raw mapped triples, and hybrid retrieved chunks.   \\ \hline
			$f$ & The mapped triplet fact in KG.  \\ \hline
			$ \mathcal{F}_{\mathcal{KG}}$ & All triplet facts within the KG. \\ \hline
			$\mathcal{F}_{raw}$, $\mathcal{F}$ & Raw and mapped triplet facts after filter.  \\ \hline
			$v$ & A node (entity) in KG.\\ \hline
			$e_{(u,v)}$ & An edge (relation) between node $u$ and $v$.  \\ 		\hline
			
			$\mathcal{KG}(\mathcal{V},\mathcal{E})$ & KG with entity set  $\mathcal{V}$ and relation set $\mathcal{E}. $ \\ \hline
			
			$d$, $\mathbf{p}$, $\mathbf{P}$ & The damping factor, personalization vector and transition probability matrix of PPR. \\ \hline
			
			$\mathcal{N}(\cdot)$& 1-hop neighbor set of node or subgraph in KG.	\\ 		\hline		
			$r_{MCMI}$ & MCMI subgraph's total cost-score ratio. \\ \hline
			$g$ & MCMI subgraph's text-formed graph string. \\ \hline
			$p_{re}$, $p_{tf}$, $p_{gen}$& LLM prompt for relation extraction, triplet filtering, and answer generation.	\\ \hline					
			% $p_{gen}$& The LLM prompt for  answer generation. 	\\ 	\hline		
			% $p_{classify}$&The classification instruction prompt. 	\\ 	\hline		
			$AS(\cdot)$, $HS(\cdot)$ & The anchor score and hybrid retrieval score.\\ 	\hline
		 $EC(\cdot)$ & The edge cost function.\\ 	 
			
			\hline\hline
		\end{tabular}
	}
	\vspace{-10px}
\end{table}
% \linespread{1.1}

\subsubsection{\textbf{Walking-based Models}}
To further enhance the long-range graph reasoning capabilities of community-based models, researchers have proposed walking-based models \cite{gutierrez2024hipporag,fastgraphrag,listructrag,gutiérrez2025ragmemory}, which apply walking-based graph algorithms, to detect and capture global topological information within the Knowledge Graphs (KGs).

Among these, HippoRAG \cite{gutierrez2024hipporag} and Fast-GraphRAG \cite{fastgraphrag} utilize the Personalized PageRank (PPR) algorithm to traverse the constructed KG and assign relevance scores to graph nodes with respect to the input query. During retrieval, these models select text chunks based on the aggregated PPR scores of the graph nodes they contain. StructRAG \cite{listructrag} trains a router module that learns to identify the optimal reasoning path over the graph based on the input query. HippoRAG2 \cite{gutiérrez2025ragmemory} further improves upon HippoRAG by incorporating text chunks as passage nodes directly into the KG. It retrieves text chunks based on the individual PPR scores of their corresponding passage nodes within the graph.

In this paper, we propose a novel graph-based and efficient RAG framework, termed AGRAG. AGRAG first constructs a KG using a statistics-based entity extraction method, and then assigns graph's node weights based on both the KG structure and the input query. Finally, it generates a Minimum Cost and Maximum Influence (MCMI) reasoning subgraph to provide the LLM with explicit reasoning chains that maximize query-relevant information while minimizing noise during retrieval.

% \vspace{-8px}
\section{Methodology}\label{sec:methdology}
% {\color{blue}
In this section, we first introduce the overall AGRAG framework, then detail its three core steps: Data Preparation, Graph Retrieval, and Hybrid Retrieval. 
% Finally, we discuss the computational hardness and complexity analysis of AGRAG.
% }
\begin{figure*}[t]
	\centering
	% \vspace{-1em}
	\includegraphics[width=\linewidth]{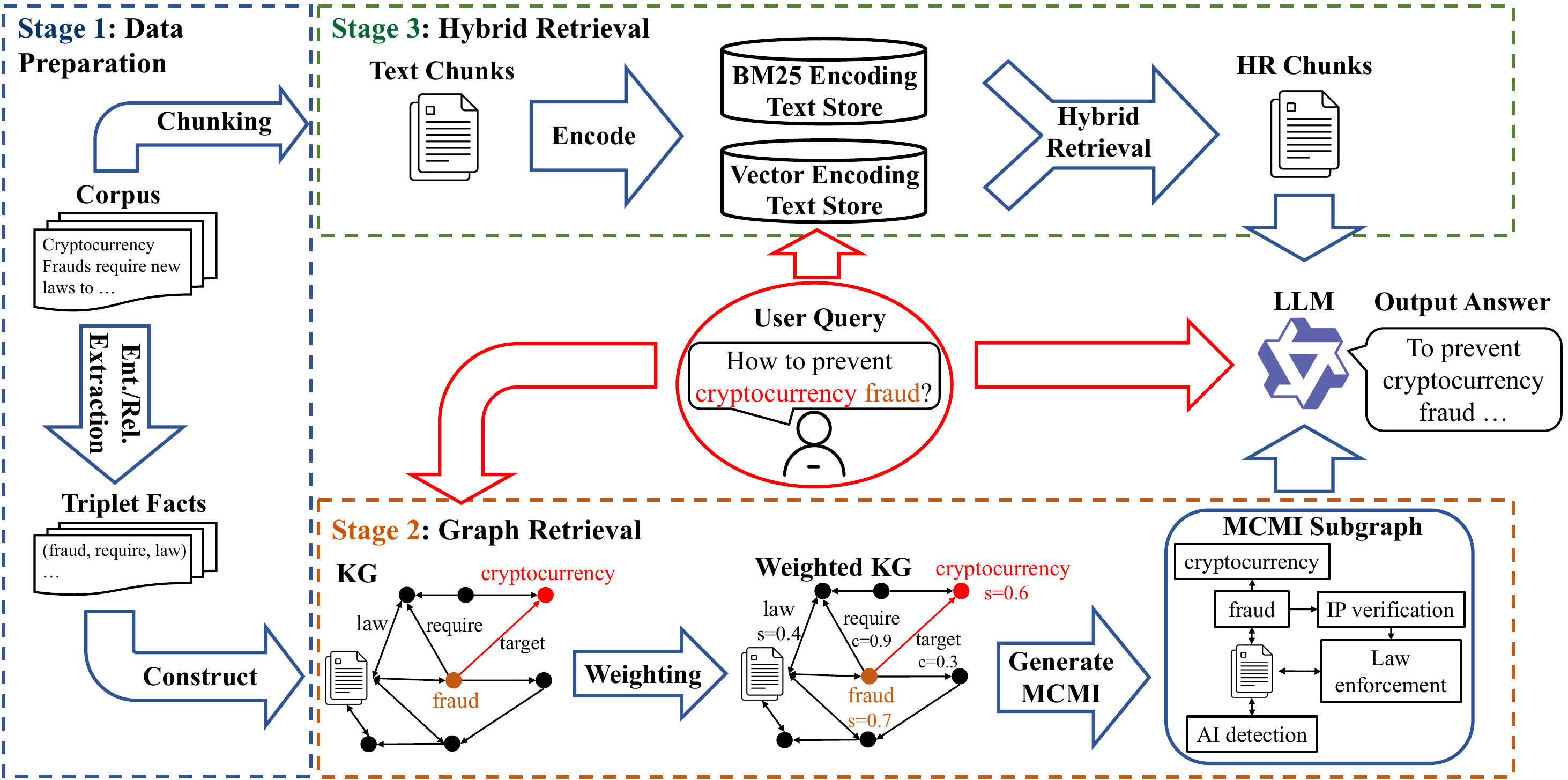}
	% \vspace{-20px}
	\caption{An overview of AGRAG. 
		In {\color{myblue}Stage 1}, AGRAG constructs a KG 
		% with a statistics-based entity extraction method 
		based on the text corpus. 
		In {\color{myorange}Stage 2}, the query is mapped to {\color{red}relevant triplet facts} in KG, then node score and edge weights are assigned based on PPR algorithm and the semantic similarity between query and each fact, respectively.
		In {\color{mygreen}Stage 3}, AGRAG generates an MCMI subgraph, providing explicit reasoning chains in addition to the most semantically similar text chunks by Hybrid Retrieval (HR) to support LLM query answering.
	}
	\label{fig:Gorag}
	\vspace{-10px}
\end{figure*}

\subsection{Framework Overview}
% \footnote{\# too redundant overview. You only need to introduce the basic idea of method and corresponding advantage. No need to introduce the details}
% \footnote{\# wrong information flows in Figure 2}
%{\color{red}
	 {
	 	% \color{blue}
% To address the aforementioned poor reasoning ability, high computational cost, and non-generalizable issue of current Graph-based RAG models. 
% We propose AGRAG, an efficient and generalizable approach by applying a statistics-based method to substitute the LLM entity extraction approach, and generate query-aware reasoning subgraphs to aid LLM inference without the need for any parameter finetuning. 
As shown in Figure \ref{fig:Gorag}, AGRAG consists of three steps, i.e., Data Preparation, Graph Retrieval, and Hybrid Retrieval.

\noindent\textbf{Step 1: Data Preparation. } 
In this step, AGRAG constructs a Knowledge Graph (KG) $\mathcal{KG}(\mathcal{V},\mathcal{E})$ based on the given text corpus $\mathcal{T}$.
% and split the texts in $\mathcal{T}$ into text chunks. 
Here, we substitute other Graph-based RAG baselines' LLM entity extraction to a statistics-based method based on TFIDF \cite{tfidf}, and extract entities from text chunks in $\mathcal{T}_c$ that split from the texts in corpus $\mathcal{T}$. 
This method can avoid LLM hallucination, and hence reduce the noise in the constructed graph, leading to better effectiveness.
After entity extraction, we apply LLM to detect relation edges w.r.t. the extracted entity nodes and the text chunk. 
% The process of this step is similar to other Graph-based RAG baselines. 
% Firstly, we chunk texts in corpus $\mathcal{T}$ into chunks $\mathcal{T}_c$.
	% \footnote{\# haoyang: is it framework overview? Yubo: Modified, each statge consists no more than 15 lines, as in Fight with Fire.}
% Next, we replace other baselines' clumsy LLM entity extraction procedure with a simple statistics-based entity extraction approach based on TFIDF \cite{tfidf}, and carry out entity extraction based on $\mathcal{T}_c$, the experiment result in \autoref{sec:experiment} demonstrates our method resulting in better efficiency but not affecting the model effectiveness. }
% After the entity extraction procedure, we apply LLM to detect relation edges between entity extraction extracted entity nodes in each text chunk. 
Furthermore, we also construct text chunks into KG as passage nodes, and link them to all their extracted entity nodes with node-to-passage edges. 
The detailed pseudo code of Step 1 is shown in Algorithm \ref{alg:graph_construction}, and more details can be found in Section~\ref{sec:gc}.

\noindent\textbf{Step 2: Graph Retrieval. } 
In this step, we firstly map the query $q$ to KG fact triples considering their semantic similarities, after that, we calculate the PPR score \cite{haveliwala2002topic} based on these mapped fact triples $F$ as the node influence score $\mathbf{s}$. 
% as the PPR score already demonstrates remarkable effectiveness and efficiency over many graph-based models \cite{gutierrez2024hipporag,fastgraphrag,gutiérrez2025ragmemory}. 
On the other hand, we calculate the edge costs $\mathcal{C}_{\mathcal{E}}$ considering the semantic similarity between query and facts' vector representations. 
the higher their similarity is, the lower the cost is. 
% An example of the weighted graph created on the Medical dataset of GraphRAG-bench \cite{xiang2025use} is shown in Figure \ref{fig:ig}. 
% The graph formed like this will be used for later Minimum Cost Maximum Influence (MCMI) subgraph generation. 
Compared with solely applying PPR to score nodes for retrieval, 
% our edge cost further provides a constraint for involving nodes w.r.t. the semantic correspondence between query and facts,  
the edge cost further provides constraints on the involved nodes based on the semantic correspondence between query and facts, 
% the edge cost further captures the semantic correspondence between the query and each fact, 
removing the need for the human-defined PPR retrieval threshold.
After that, we initialize the MCMI subgraph with the generated MCST, and apply our greedy algorithm by iteratively expanding the MCMI subgraph by selecting its neighboring node with the lowest edge cost-influence score ratio (i.e., $c/s$). 
If the lowest ratio is larger than the total ratio of the current MCMI, the iteration stops and the current MCMI subgraph is returned. 
We prove the MCMI generation problem to be NP-hard, and our greedy algorithm in Algorithm \ref{alg:mehlhorn} can achieve a 2-approximation. Please refer to Section~\ref{sec:gr} for further details.
% and
% enabling the Minimum Cost Maximum Influence (MCMI) subgraph generation in Step 3. 
% Consequently, our graph weighting mechanism can provide complex structured reasoning information, 
% enabling LLMs to better understand the correspondence between retrieved text chunks and the input query. 
% The detailed pseudo code of Step 2 is shown in Algorithm Algorithm \ref{alg:graph_weighting}, and 
% More details of Step 2 can be found in Section~\ref{sec:gr}.

\noindent\textbf{Step 3: Hybrid Retrieval. } 
In this step, 
we perform hybrid retrieval and obtain text chunks based on their BM25 sparse similarity score, and dense vector representation similarity score with the query. 
These chunks will be input together with the MCMI subgraph to the LLM for the final query answering, providing extra contexts.

}

\subsection{Step 1: Data Preparation} \label{sec:gc}
% \footnote{\# the title has problems, it is not indexing Yubo: Fixed}
% \footnote{\# the motivation lacks supports. In another words, whey reviewers need to believe such simple method can work, and why you use such ways. Any clues to support your choice and method Yubo: Modified}
In this subsection, 
% we introduce the graph construction procedure of AGRAG, where 
we utilize a TFIDF \cite{tfidf} based entity extraction approach to substitute the LLM-based approach of other Graph-based RAG models, 
reduce LLM hallucination, resulting in less noise generated from the early entity extraction step, and hence less error propagation, 
address the inaccurate graph construction issue, 
leading to better effectiveness.
% address the High Computational Cost issue of current Graph-based RAG models, resulting in better efficiency but not affect the effectiveness.
The pseudo code of AGRAG's graph construction procedure is shown in Algorithm \ref{alg:graph_construction}.
\begin{algorithm}[t]
	\caption{Indexing Algorithm of AGRAG\label{alg:graph_construction}}
	\KwIn{$\mathcal{T}$: The text corpus for graph construction.}
	\KwOut{$\mathcal{KG}(\mathcal{V},\mathcal{E})$: The KG constructed.}
	Let $\mathcal{V} \gets \emptyset$, $\mathcal{E} \gets \emptyset$\;
	Get text chunk set $\mathcal{T}_c$ from $\mathcal{T}$: $\mathcal{T}_c \gets \text{Chunking}(\mathcal{T})$\;
	\For{each text chunk $t \in \mathcal{T}_c$}{
		Extract entity set $\mathcal{V}_t$ based on Equation \eqref{eq:entity_extraction}\;
		Carry out relation extraction with LLM: $\mathcal{E}_t \gets \text{LLM}(\mathcal{V}_t, t, p_{\text{re}})$\;
		$\mathcal{V} \gets \mathcal{V} \cup \mathcal{V}_t$, $\mathcal{E} \gets \mathcal{E} \cup \mathcal{E}_t$\;
	}
	Add $\mathcal{T}_c$ to $\mathcal{V}$ as passage nodes: $\mathcal{V} \gets \mathcal{V} \cup \mathcal{T}_c$\;
	\For{each entity node $v \in \mathcal{V}$}{
		\For{each text chunk $t \in \mathcal{T}_c$}{
			\If{$v \in t$}{
				Add passage edge between $v$ and $t$: $\mathcal{E} \gets \mathcal{E} \cup \{e_{(v,t)}\}$\;
			}
		}
	}
	\Return{$\mathcal{KG}(\mathcal{V},\mathcal{E})$}\;
\end{algorithm}
\begin{comment}
\begin{algorithm}[t]
	\caption{Indexing Algorithm of AGRAG}\label{alg:graph_construction}
	\begin{algorithmic}[1]
		\REQUIRE $\mathcal{T}$: The text corpus for graph costruction. 
		\ENSURE $\mathcal{KG}(\mathcal{V},\mathcal{E})$: The Knowledge Graph constructed.
		\STATE Let $\mathcal{V}=\emptyset$, $\mathcal{E}=\emptyset$
		\STATE Get text chunk set $\mathcal{T}_c$ from $T$: $\mathcal{T}_c=\text{Chunking}(\mathcal{T})$;
		\FOR{each text chunk $t\in \mathcal{T}_c$}
		\STATE Extract entity set $\mathcal{V}_t$ based on Equation \eqref{eq:entity_extraction};
		 %: $\mathcal{V}_t=entity extraction(t)$;
		\STATE Carry out relation extraction with LLM: $\mathcal{E}_t=\text{LLM}(\mathcal{V}_t,t,p_{\text{re}})$;
		% \STATE Add edges between all entity nodes in $\mathcal{V}_t$ and text chunk $t$ and obtain passage edge set $\mathcal{E}_p$;
		\STATE $\mathcal{V}=\mathcal{V}\cup \mathcal{V}_t$, $\mathcal{E}=\mathcal{E}\cup \mathcal{E}_t$;
		\ENDFOR 
		\STATE Add $\mathcal{T}_c$ to $\mathcal{V}$ as passage node: $\mathcal{V}=\mathcal{V}\cup \mathcal{T}_c$;
		\FOR{each entity node $v\in \mathcal{V}$}
		\FOR{each text chunk in $t\in \mathcal{T}_c$}
		\IF{$v\in t$}
		\STATE Add passage edge between $v$ and $t$: $\mathcal{E}=\mathcal{E}\cup e_{(v,t)}$;
		\ENDIF
		\ENDFOR
		\ENDFOR
		\RETURN  $\mathcal{KG}(\mathcal{V},\mathcal{E})$
	\end{algorithmic}
\end{algorithm}
\end{comment}
% in \autoref{sec:pseduo_code}.
% }

To begin with, we split texts in $\mathcal{T}$ into fixed-length chunks with a certain amount of token overlap between chunks: 
\begin{equation}\label{eq:chunking}
	\mathcal{T}_c=\text{Split}(\mathcal{T},l_t,l_o),
\end{equation}
where $l_t$ and $l_o$ denotes the length of text chunks and token overlap between chunks. 
{
	% \color{blue}
Next, we apply the statistics-based entity extraction approach on the text chunks $\mathcal{T}_c$. 
Specifically, our approach 
% calculates the term frequency score, 
modifies the TFIDF score \cite{tfidf} as the entity extraction score $\text{ER}(v,t)$ for each term $v$ in text chunk $t$ as an entity, to make the value of ER fall in $(0,1)$. 
Here, our entity detection is based on an n-gram setting with the maximum n-gram (number of words in an entity) as $b$, and the equation of $\text{ER}(v,t)$ is as follows:
% Different from the original TFIDF, to avoid high compotational cost on large scale of corpus, we calculate the IDF score only in a: 
% \footnote{so where is your contributions. The writing style directly told to reviewers: we just use tf-idf, and no contributions Yubo: Modified writing style.}
}
\begin{equation}\label{eq:tfidf_ke}
	% \vspace{-5px}
	\text{ER}(v,t)=\frac{\text{count}(v,t)}{|t|log(|\mathcal{T}_c|+1)}\times log\frac{|\mathcal{T}_c|+1}{|t_j:v\in t_j, \ t_j \in \mathcal{T}_c|+1}, 
\end{equation}
where $ \mathcal{T}_c$ denotes all text chunks obtained from Equation \eqref{eq:chunking}, 
$\text{count}(v, t)$ is the number of times that the term $v$ appears in the text $t$, 
and $|t_j : v \in t_j, \ t_j \in \mathcal{T}_c|$ denotes the number of texts in the corpus $\mathcal{T}_c$ that contain the keyword $v$. 
Here, our entity extraction procedure will extract words or phrases with $\text{ER}(v,t)>\tau$ as the entity nodes $\mathcal{V}_t$ for text chunk $t$:  
% These entities are served as graph nodes, hence we denote them as entity nodes. 
\begin{equation}\label{eq:entity_extraction}
	\mathcal{V}_t = \left\{ v \,\middle|\, \text{ER}(v, t) > \tau \right\}.
\end{equation}
Entities in $\mathcal{V}_t$ are later served as graph nodes, hence we denote them as entity nodes. 
For our entity extraction procedure, the entity boundary $b$ and entity extraction threshold $\tau$ are considered as hyper-parameters, we further analyze their setting in Section \ref{sec:para_sen}.

After entity extraction, we extract relations from text chunks $t\in\mathcal{T}_c$ based on its extracted entities $\mathcal{V}_t$. Align with other Graph-based RAG models, we also apply LLM at this step:

\begin{equation}
	\label{eq:relation_ext}
	\mathcal{E}_t=\text{LLM}(\mathcal{V}_t,\ t,\ p_{\text{re}}), 
\end{equation}
where $\mathcal{V}_t$ denotes the extracted entities of text chunk $t$, and in the relation extraction prompt $p_{\text{re}}$, we ask the LLM to detect relations between entities in $\mathcal{V}_t$ w.r.t. text chunk $t$. 
Although we also apply LLM for relation extraction, we input less noise in this step, mitigating the error propagation, as our statistics-based entity extraction avoids LLM hallucination, hence resulting in a more comprehensive and less noisy $\mathcal{V}_t$ as the Equation \eqref{eq:relation_ext} input.

After relation extraction, we add text chunks in $\mathcal{T}_c$ to $\mathcal{V}$ as passage nodes, and add passage edges $e_{(v,t)}$ between entities contained within the text chunks to $\mathcal{E}$, the respective relation of these edges is namely \textit{Contains}. 
By introducing passage nodes, we integrate contextual information into the constructed KG, and the entity-passage edges can capture  the inter-text correlations, make our model able to generate reasoning paths across different text chunks \cite{gutiérrez2025ragmemory}.

Furthermore,  we add synonym edges $\mathcal{E}_\text{syn}$ between entity nodes in $\mathcal{V}$ that have similarities larger than the synonym threshold $\phi$, the relation of these edges is namely \textit{synonym}: 
\begin{equation}
	\mathcal{E}_\text{syn} = \left\{ e_{(u,v)}\,\middle|\, \text{cosine\_sim}(\textbf{u},\textbf{v})>\phi \right\},
\end{equation}

At last, we combine edge set $\mathcal{E}_t$ for all text chunks $t$ and passage edges together, as well as the synonym edges $\mathcal{E}_\text{syn}$, to form the total edge (relation) set of the KG:
\begin{equation}
\mathcal{E}=\bigcup_{t\in \mathcal{T}_c}\mathcal{E}_t \ \cup \bigcup_{v\in \mathcal{V}_t,\ t\in \mathcal{T}_c} e_{(v,t)} \ \cup \mathcal{E}_\text{syn},
\end{equation}
here, if the entity $v$ does not exist in text chunk $t$, $e_{(v,t)}=\emptyset$.

\begin{figure}[t]
	\centering
	\includegraphics[width=1\linewidth]{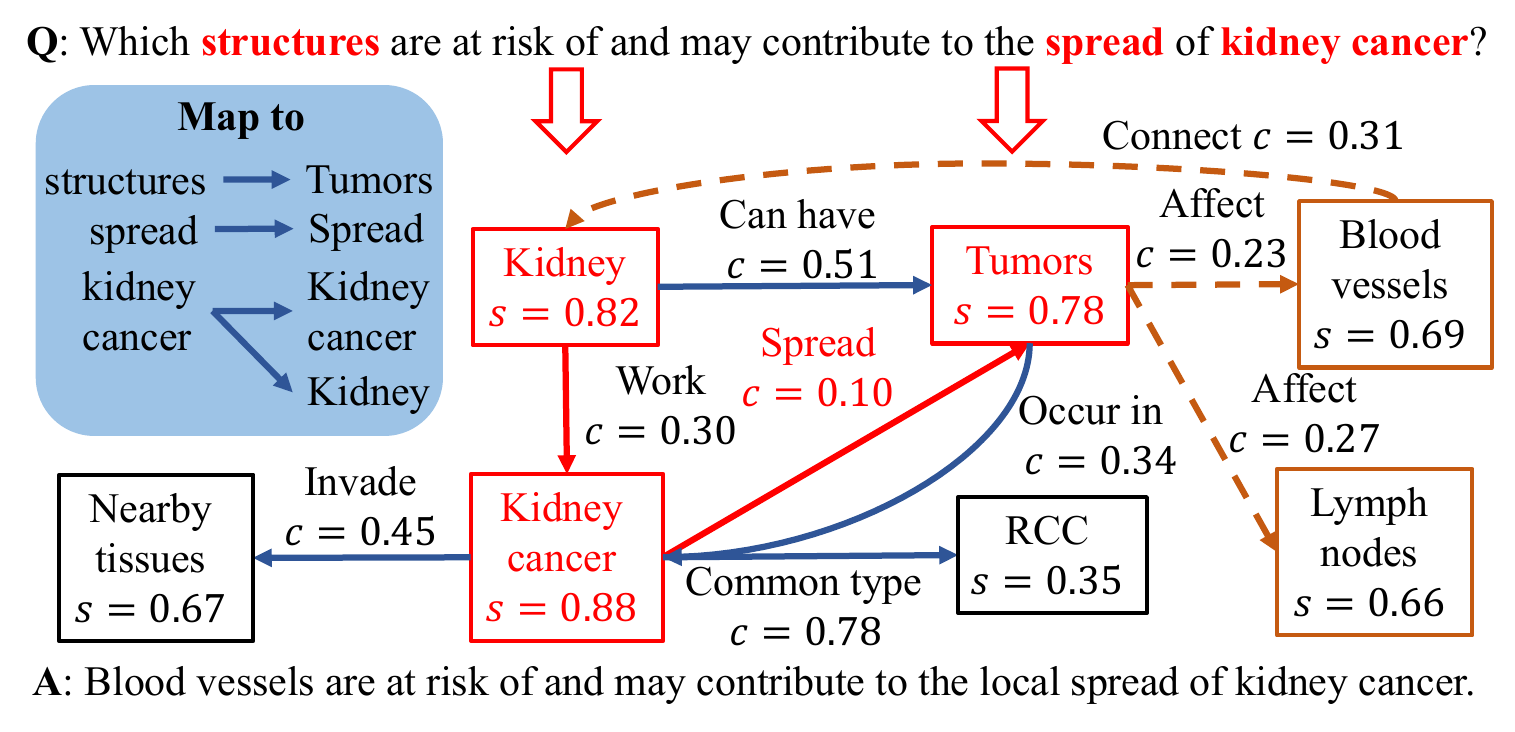}
	% \vspace{-20px}
	\caption{An example of MCMI. 
		% on our weighted KG. 
		Red triples indicate those mapped from the query. Our algorithm first constructs the MCMI based on a Steiner tree rooted at these mapped triples, then iteratively expands it by adding neighboring triples (in {\color{myorange} orange}) with the lowest cost-score ratio, continuing until no neighboring triple has its ratio less than the current MCMI's ratio. 
	}
	\label{fig:ig}
	\vspace{-10px}
\end{figure}

% \vspace{-9px}
\subsection{Step 2: Graph Retrieval}\label{sec:gr}
% \vspace{-1px}
In this subsection, we introduce the graph retrieval procedure of AGRAG. 
Given the KG $\mathcal{KG}(\mathcal{V},\mathcal{E})$ constructed at the Step 1, we first map the query $q$ to $\mathcal{KG}$ and obtain raw mapped triplet facts $\mathcal{F}_\text{raw}$. 
To achieve this, we encode each triplet fact in KG $f\in\mathcal{F}_{\mathcal{KG}}$ and the query $q$ into vector representations $\textbf{q}$ and $\textbf{f}$ with encoder $\mathcal{M}$: $\textbf{q} \gets \mathcal{M}(q)$, then select facts with top-k dense semantic mapping score MS to query $q$ based on their cosine similarity: 
\begin{equation}
	\label{eq:cos_sim}
	\text{MS}(\textbf{q},\textbf{f})=\text{cosine\_sim}(\textbf{q},\textbf{f}),
\end{equation}
\begin{equation}
	\label{eq:top_k}
	\mathcal{F}_\text{raw} = \left\{ f \in \mathcal{F}_{\mathcal{KG}}\,\middle|\, f \in \arg\top_{k_a} \left(MS(\textbf{q},\textbf{f}) \right) \right\},
\end{equation}
where $k_a$ is a hyperparameter that controls the number of raw mapped triples extracted, and $\textbf{f}$ are obtained through encoding the text concatenation of entities and relation of the triplet fact. 

% \textcolor{blue}{
	Further, as the triplets in KG are extracted in a query unaware manner, query irrelevant triplets can exist in the raw mapped triplet facts $\mathcal{F}_\text{raw}$, hence we apply LLM-based filter on $\mathcal{F}_\text{raw}$ to generate the mapped triplet facts $\mathcal{F}$, to reduce the query irrelevant noise:
	% }:
\begin{equation}
	\label{eq:triplet_filter}
	\mathcal{F}=\text{LLM}(\mathcal{F}_\text{raw}, q, p_\text{tf}),
\end{equation}
where $p_\text{tf}$ denotes the LLM prompt for triple filtering. 
% in $p_\text{tf}$, we ask the LLM to filter out unrelated triplets w.r.t. query $q$.

Next, based on the mapped triplet facts $\mathcal{F}$ and the KG structure, we calculate the Personalized PageRank (PPR) \cite{haveliwala2002topic} score as our node influence score $\mathbf{s}$ for node set $\mathcal{V}$: 
\begin{equation}\label{eq:ppr}
	\mathbf{s} = \lim_{i \to \infty} \text{PPR}^i, \, \text{PPR}^i=(1-d)\times \mathbf{p}+d\times \mathbf{P} \cdot \text{PPR}^{i-1}, 
\end{equation}
and we have the node score set $\mathcal{S}_{\mathcal{V}}=\{\mathbf{s}[v] | v\in\mathcal{V}\}$.
 % To align with the settings in \cite{gutiérrez2025ragmemory}, when the PPR score changes less than $10^{-7}$, the calculations are considered converged, and we consider the score at this moment
To align with \cite{gutiérrez2025ragmemory}, we set the damping factor $d$ to 0.5, and converge the PPR calculation when it changes less than $10^{-7}$; 
% where $d$ denotes the damping factor, we set it to 0.5 to align with the settings in \cite{gutiérrez2025ragmemory}, 
% {\color{blue}
$\mathbf{p}\in \mathbb{R}^{|\mathcal{V}|}$ denotes the personalization vector that depends on the mapped triples $\mathcal{F}$:
\begin{equation}
	\label{eq:pers_vector}
	\mathbf{p}[u]=\begin{cases} 
		1, & \text{if } u \in \mathcal{F}, \\
		0, & \text{otherwise}.
	\end{cases}
\end{equation}
% where the value of corresponding elements of the mapped triples' entities are set to $1/|\mathcal{F}|$, and other elements are set to $0$. 
The $\mathbf{P}\in \mathbb{R}^{|\mathcal{V}\times \mathcal{V}|}$ in Equation \eqref{eq:ppr} denotes the transition probability matrix, and can be obtained by: 
\begin{equation}
	\label{eq:tp_matrix}
	\mathbf{P}[u][v] = 
	\begin{cases} 
		\frac{1}{|\mathcal{N}(u)|}, & \text{if } e_{(u,v)} \in \mathcal{E}, \\
		\frac{1}{|\mathcal{V}|}, & \text{otherwise},
	\end{cases}
\end{equation}
where $\mathcal{N}(u)$ denotes the 1-hop neighbor set of node $u$ in the KG. 
% \textcolor{blue}{
	To ensure the PPR's faster convergence and more stable iterative computation \cite{boldi2005pagerank}, as well as normalizing the score distribution \cite{baeza2006generic}. 
	Following \cite{gutiérrez2025ragmemory}, we set the transition probability for non-edges to $\frac{1}{|\mathcal{V}|}$, and after calculating the PPR score, we multiply the PPR score of all passage nodes by a balance factor 0.05, to avoid over concentration on them. % }
After that, we can obtain the node influence score $s_v$ for each node $v\in \mathcal{V}$ as follows:
\begin{equation}\label{eq:node_score}
	s_v = \mathbf{s}[v], \quad \forall v \in \mathcal{V}.
\end{equation}

\begin{algorithm}[t!]
	\caption{Graph Weighting Algorithm\label{alg:graph_weighting}}
	\KwIn{
		$\mathcal{KG} (\mathcal{V}, \mathcal{E})$: The indexed KG; \\
		$\mathcal{F}$: A set of query-mapped triplet facts in $\mathcal{KG}$; \\
		$\mathbf{q}$: The vector representation of the query $q$.
	}
	\KwOut{
		$\mathcal{KG} (\mathcal{V}, \mathcal{E}, \mathcal{S}_\mathcal{V}, \mathcal{C}_\mathcal{E})$: The weighted KG.
	}
	Assign node weights $\mathcal{S}_\mathcal{V}$ with PPR score w.r.t. $\mathcal{F}$ based on Equation \eqref{eq:ppr}: $\mathbf{s} \gets \lim_{i \to \infty} \text{PPR}^i$\;
	\For{$e_{(u,v)} \in \mathcal{E}$}{
		Assign edge costs $c_{e_{(u,v)}}$ based on Equation \eqref{eq:edge_weigting}: $c_{e_{(u,v)}} \gets (1 - \text{MS}(\mathbf{q}, \mathbf{f}_{e_{(u,v)}})) / 2$\;
		$\mathcal{C}_\mathcal{E} \gets \mathcal{C}_\mathcal{E} \cup \{c_{e_{(u,v)}}\}$\;
	}
	\Return{$\mathcal{KG} (\mathcal{V}, \mathcal{E}, \mathcal{S}_\mathcal{V}, \mathcal{C}_\mathcal{E})$}\;
\end{algorithm}

\begin{comment}
\begin{algorithm}[t!]
	\caption{Graph Weighting Algorithm}
	\label{alg:graph_weighting}
	\begin{algorithmic}[1]
		\REQUIRE $\mathcal{KG} (\mathcal{V}, \mathcal{E})$: The indexed KG;
		
		$\mathcal{F}$: A set of query mapped triplet fact in $\mathcal{KG}$;
		
		$\textbf{q}$: The vector representation of the query $q$.
		\ENSURE $\mathcal{KG} (\mathcal{V}, \mathcal{E},\mathcal{S}_\mathcal{V},\mathcal{C}_\mathcal{E})$: The weighted KG.
		\STATE Assign node weights $\mathcal{S}_\mathcal{V}$ with PPR score w.r.t. $\mathcal{F}$ based on Equation \eqref{eq:ppr}: $\mathbf{s} = \lim_{i \to \infty} \text{PPR}^i$;
		\FOR{$e_{(u,v)} \in \mathcal{E}$}
		\STATE Assign edge costs $c_{e_{(u,v)}}$ based on Equation \eqref{eq:edge_weigting}: $c_{e_{(u,v)}}=(1 - \text{MS}(\mathbf{q}, \mathbf{f}_{e_{(u,v)}}))/2$;
		\STATE $\mathcal{C}_\mathcal{E}=\mathcal{C}_\mathcal{E}\cup c_{e_{(u,v)}}$;
		\ENDFOR
		\RETURN $\mathcal{KG} (\mathcal{V}, \mathcal{E},\mathcal{S}_\mathcal{V},\mathcal{C}_\mathcal{E})$
	\end{algorithmic}
\end{algorithm}
\end{comment}
At last, we calculate the edge cost $\mathcal{C}_{\mathcal{E}}$ based on the similarity between query $q$'s representation $\textbf{q}$ and relation edge $e_{(u,v)}$'s respective triplet facts $f_{e_{(u,v)}}$'s representation $\textbf{f}_{e_{(u,v)}}$: 

\begin{equation}\label{eq:edge_weigting}
	\mathcal{C}_{\mathcal{E}} = \{ c_{e_{(u,v)}} \mid e_{(u,v)} \in \mathcal{E} \}, \quad 
	c_{e_{(u,v)}} = \frac{1 - \text{MS}(\mathbf{q}, \mathbf{f}_{e_{(u,v)}})}{2}.
\end{equation}

Moreover, to guarantee the KG's connectivity, we connect all passage nodes to a pseudo node, with the pseudo edge (relation) namely \textit{pseudo\_relation}, and the cost of these pseudo edges is set to 10 to penalize paths through the pseudo node, aligning with the setting of HippoRAG2 \cite{gutiérrez2025ragmemory}.

Next, we introduce the Minimum Cost Maximum Influence (MCMI) subgraph generation problem, which 
% on the weighted KG to 
can provide more comprehensive reasoning chains and result in more comprehensive answers in tasks that require knowledge summarization. 
According to Figure \ref{fig:ig}, the MCMI subgraph can include cyclic graph structures to cover key query related triples (colored in {\color{myorange}orange}), which are not covered by simple node matching or Steiner tree, hence address the Poor Reasoning Ability and Inadequate Answer issue of current Graph-based RAG models. 
Here, we give a formal definition of the MCMI subgraph generation problem: 
% {\color{blue}
\begin{definition} [Minimum Cost Maximum Influence Subgraph Generation Problem] 
\label{def:MCMI} 
Given a node-weighted and edge-weighted, connected graph $\mathcal{G}(\mathcal{V},\mathcal{E},\mathcal{S}_\mathcal{V},\mathcal{C}_\mathcal{E})$ and a set of mapped triples $\mathcal{F}$,
let $\mathcal{V}_{\text{term}}$ be the set of nodes in $\mathcal{F}$, called terminals. 
Let $r\in\mathcal{V}_{\text{term}}$
be a designated root terminal.
The goal is to generate a subgraph $\mathcal{G}_\text{MCMI}(\mathcal{V}',\mathcal{E}', \mathcal{S}'_\mathcal{V}, \mathcal{C}'_\mathcal{E})$
that contains all terminals $\mathcal{V}_{\text{term}}$ such that every terminal
$v\in \mathcal{V}_{\text{term}}$ is reachable from root terminal $r$, 
% (i.e., there exists a path from $r$ to $v$ within the subgraph),
while minimizing the total cost-score ratio $r(\mathcal{G_{\text{MCMI}}})$:
\begin{equation}\label{eq:r_mcmi}
	r(\mathcal{G}_{\text{MCMI}})=\sum_{e_{(u,v)}\in \mathcal{E}'} \frac{c_{e_{(u, v)}}}{s_{u}+s_v},
\end{equation}
where	$s_v = \mathbf{s}[v]$ denotes the Personalized PageRank (PPR) score of node $v$, as defined in Equation~\eqref{eq:ppr}, and $c_{e_{(u, v)}}$ represents the edge cost of $e_{(u, v)}$, as defined in Equation~\eqref{eq:edge_weigting}.
% average node score and minimizing the total edge cost.
\end{definition}

By minimizing the cost-score ratio in Equation \eqref{eq:r_mcmi}, we can ensure the generated subgraph maximizes the inclusion of high-influence, query-relevant nodes while strictly penalizing low-similarity connections to effectively filter out noise.

% At first, we give the definition of the MCS problem.

% \begin{definition} [Minimum Set Cover Problem] 
% 	Given a universe $\mathcal{U}$ and a subset family $\mathcal{SF}\subseteq 2^{\mathcal{U}}$,
% 	find a minimum-cardinality subfamily (minimum-set cover) $\mathcal{CSF}\subseteq \mathcal{SF}$ such that the union of all subet $s\in\mathcal{CSF}$ satisifies 
% 	$\bigcup_{s\in \mathcal{CSF}} s = \mathcal{U}$.
% \end{definition}

\begin{theorem}\label{thm:unboundedness}
	The MCMI subgraph generation   is NP-hard.
\end{theorem}

\begin{proof}[Proof Sketch]
	We establish NP-hardness by constructing a bijection between the MCMI problem and the classic Steiner Tree problem via a transformed weight function. Due to the space limitation, we put our proof at \url{https://github.com/Wyb0627/AGRAG/blob/main/proof.pdf}.
\end{proof}

Since the MCMI Subgraph Generation problem is NP-hard, it is infeasible to obtain the optimal result in polynomial time.
Therefore, we propose a greedy algorithm (Algorithm \ref{alg:mehlhorn}) based on the Mehlhorn's algorithm \cite{mehlhorn1988faster} to solve this problem. 
% Then, we apply our greedy algorithm by iteratively expanding the generated MCMI. 
% Here, we first 

To begin with, we initialize the MCMI as the Minimum Cost Steiner Tree (MCST) through all mapped triplet facts $\mathcal{F}$ in the weighted graph $\mathcal{G} (\mathcal{V}, \mathcal{E},\mathcal{S}_\mathcal{V},\mathcal{C}_\mathcal{E})$ (line 1 to line 10 of Algorithm \ref{alg:mehlhorn}), where score $s_v\in\mathcal{S}_\mathcal{V}$ of each node $v$, and cost $c_{e_{(u,v)}}\in\mathcal{C}_\mathcal{E}$ of each edge $e$ are obtained with Equation \eqref{eq:node_score} and Equation \eqref{eq:edge_weigting}, respectively:
\begin{equation}
	\mathcal{G}_{\text{MCMI}}(\mathcal{V}_{\text{MCMI}},\mathcal{E}_{\text{MCMI}})=\text{MCST}(\mathcal{F},\mathcal{G}),
\end{equation}
where $\text{MCST}(\mathcal{F},\mathcal{G})$ denotes the Minimum Cost Steiner tree generation algorithm.

Next, we calculate the ratio of the edges' cost and their connected nodes' influence score (i.e., $c/s$), and check this ratio of all nodes in the current MCMI's 1-hop neighborhood $\mathcal{N}(\mathcal{G}_\text{MCMI})$, if this ratio is less than that of the current MCMI subgraph's total cost-score ratio, we add this node (entity) and the corresponding edge (relation) to the MCMI, recalculate the MCMI's cost-score ratio and continue the iteration, until all neighbor node's cost-score ratio is more than or equal to the MCMI subgraph's total cost-score ratio. That is, for each edge $e_{(u,v)}$, we have (line 10 to line 16 of Algorithm \ref{alg:mehlhorn}): 
\begin{equation}\label{eq:MCMI}
	\begin{aligned}
		\mathcal{V}_\text{MCMI}&= 
		\begin{cases} 
			\mathcal{V}_\text{MCMI} \cup v, & \text{if } \frac{c_{e_{(\text{MCMI}, v)}}}{s_v}<r(\mathcal{G}_{\text{MCMI}}), \\
			\mathcal{V}_\text{MCMI}, & \text{otherwise},
		\end{cases}\\
		\mathcal{E}_\text{MCMI}&= 
		\begin{cases} 
			\mathcal{E}_\text{MCMI} \cup e_{(\text{MCMI},v)}, & \text{if } \frac{c_{e_{(\text{MCMI}, v)}}}{s_v}<r(\mathcal{G}_{\text{MCMI}}), \\
			\mathcal{E}_\text{MCMI}, & \text{otherwise},
		\end{cases}
	\end{aligned}
\end{equation}
where 
% $r(\mathcal{G}_{\text{MCMI}})$ denotes the average cost-score ratio of the current MCMI subgraph; 
$\mathcal{V}_\text{MCMI}$ and $\mathcal{E}_\text{MCMI}$ denote the node and edge set of current MCMI subgraph; 
$e_{(\text{MCMI},v)}$ denotes the outgoing edges from the MCMI subgraph to its 1-hop neighborhood. 
%And $r(\mathcal{G}_{\text{MCMI}})$ is calculated as follow:
%\begin{equation}\label{eq:r_mcmi}
%	r(\mathcal{G}_{\text{MCMI}})=\sum_{e_{(u,v)}\in \mathcal{E}_\text{MCMI}} \frac{c_{e_{(u, v)}}}{s_{u}+s_v}.
%\end{equation}

After the MCMI subgraph generation, we serialize the generated MCMI subgraph into a text graph string $g$: 
\begin{equation}
	g=\text{StrConcat}(\mathcal{V}_\text{MCMI}, \mathcal{E}_\text{MCMI}).  
\end{equation}

\noindent\textit{\textbf{Running Example:}} 
As shown in Figure \ref{fig:ig}, if we simply calculate the minimum cost Steiner tree within the given KG, the algorithm will simply return the two mapped fact (\textit{Kidney, Work, Kidney cancer}), and (\textit{Kidney, Spread, Tumors}) as the reasoning path, and the LLM's response answer would likely be \textit{Tumors}. 
However, if we go one step further, then within the generated cyclic graph, we can have more complete reasoning information regarding the query, and involve the correct answer \textit{Blood vessels}.

We further prove our MCMI generation algorithm's MCST generation step (Lines 1-10 in Algorithm \ref{alg:mehlhorn}) can still have a 2-approximation rate in Theorem \ref{thm:MCST}.  And its MCMI generation step (Lines 11-15 in Algorithm \ref{alg:mehlhorn}) strictly monotonically gains information and terminates within at most $|\Omega|$ iterations in Theorem \ref{thm:MCMI}, where $\Omega$ denotes the candidate edge set of MCMI.
Together, they establish the effectiveness and efficiency of our MCMI generation Algorithm \ref{alg:mehlhorn}.
\begin{theorem}
	\label{thm:MCST}
	The MCST generation step (Lines 1-10 in Algorithm \ref{alg:mehlhorn}) guarantees a 2-approximation. 
\end{theorem}

\begin{proof}[Proof Sketch]
	We bound the cost of the generated subgraph by comparing the Minimum Spanning Tree (MST) on the auxiliary graph $H$ to a shortcutted Euler tour (by depth-first traversal) of the optimal Steiner tree. 
	%By invoking the triangle inequality, we show this tour has a cost of at most twice the optimal solution. 
	Due to space limitation, we put our proof at \url{https://github.com/Wyb0627/AGRAG/blob/main/proof.pdf}.
\end{proof}
\begin{theorem}
	\label{thm:MCMI}
	The greedy expansion of the MCMI generation step (Lines 11-15 in Algorithm \ref{alg:mehlhorn}) strictly monotonically gains information and terminates within at most $|\Omega|$ iterations, where $\Omega$ denotes the candidate edge set of MCMI:
	% The MCMI Subgraph Generation (Algorithm 3) guarantees a 2-approximation ratio for the connectivity cost of the mandatory reasoning backbone (Lines 1-9). And the subsequent % greedy expansion (Lines 11-15) 
	% strictly monotonically improves the subgraph's average density and 
	% converges to a local stationary point, 
	%where number of iterations 
	%$k$ 
	% is strictly bounded by the size $|\Omega|$ of the \textit{Effective Candidate Set} $\Omega$:
	\begin{equation}
		\label{eq:ecs}
		% k \le |\Omega|,\ \text{where}\ 
		\Omega = \left\{ e_{(u,v)} \in \mathcal{E} \setminus \mathcal{E}_\text{MCST} \mid \frac{c_{e_{(u,v)}}}{s_v} < r(\mathcal{G}_\text{MCST}) \right\}
	\end{equation}
\end{theorem}

\begin{proof}[Proof Sketch] We prove the MCMI generation steps' strict monotonic improvement by showing the algorithm only adds edges if their marginal cost-score ratio is strictly lower than the current subgraph's ratio. Termination is guaranteed because the expansion process is restricted to a finite set of candidate edges. Due to the space limitation, we put our proof at \url{https://github.com/Wyb0627/AGRAG/blob/main/proof.pdf}.
\end{proof}

% }
\begin{comment}
{\color{orange}
	\footnote{\color{orange} Old proof}
\begin{theorem}
	The approximation ratio of the MCMI Subgraph Generation Algorithm (Algorithm \ref{alg:mehlhorn}) is 2.
\end{theorem}
%	In Algorithm Algorithm \ref{alg:mehlhorn}, we compute the minimum cost spanning tree which connects all nodes in $\mathcal{F}$ as the Steiner tree with minimum cost, following triangle inequality, the approximation ratio of Algorithm Algorithm \ref{alg:mehlhorn} is 2. 
%	
\begin{proof}
	% 	{\color{blue}
		In Algorithm \ref{alg:mehlhorn}, we firstly compute the minimum cost Steiner tree $\mathcal{G}(\mathcal{V}_{\text{MCST}}, \mathcal{E}_{\text{MCST}})$ which connects all nodes in $\mathcal{V}_{\text{term}}$. Suppose the total cost for $\mathcal{G}$ is $\gamma$, and the optimal minimum cost Steiner tree is $\mathcal{G}^{*}(\mathcal{V}_{\text{MCST}^*}, \mathcal{E}_{\text{MCST}^*})$ with cost $\gamma^*$. $\mathcal{G}^{*}$ covers at least all edges from $\mathcal{G}$. Following triangle inequality, for $\forall e \in  \mathcal{E}_{\text{MCST}^*}$, the sum of weight is less than 2 times of edge weights in $\mathcal{E}_{\text{MCST}}$, which is $\gamma \leq 2\gamma^*$. The process from line 12 to line 15 will retrieve more edges. However, they will not change the weight in $\mathcal{G}$. Hence, the approximation ratio of Algorithm \ref{alg:mehlhorn} is 2.
		% }	
\end{proof}	
}
\end{comment}

\begin{algorithm}[t!]
	\caption{MCMI Subgraph Generation Algorithm\label{alg:mehlhorn}}
	\KwIn{
		$\mathcal{KG} (\mathcal{V}, \mathcal{E}, \mathcal{S}_\mathcal{V}, \mathcal{C}_\mathcal{E})$: The weighted KG; \\
		$\mathcal{F}$: A set of query-mapped triplet facts in $\mathcal{KG}$.
	}
	\KwOut{
		$\mathcal{G}_\text{MCMI}(\mathcal{V}_\text{MCMI}, \mathcal{E}_\text{MCMI})$: The MCMI subgraph.
	}
	% \textcolor{blue}{
		Compute an approximated minimum cost Steiner tree (MCST) $\mathcal{G}$ of $\mathcal{KG}$\;
		% }
	Let $\mathcal{V}_{\text{term}} \gets \mathcal{F} \cap \mathcal{V}_\text{MCST}$ be the terminal nodes in MCST\;
	Construct weighted auxiliary graph $H$: $\mathcal{V}_H \gets \mathcal{V}_{\text{term}}$\;
	\For{each pair of terminal nodes $u, v \in \mathcal{V}_{\text{term}}$}{
		Find the shortest path $P_{(u,v)}$ in $\mathcal{G}$  from $u$ to $v$\;
		Let $c^H_{(u, v)} \gets \sum_{e \in P_{(u,v)}} c_e$, where $c_e \in \mathcal{C}_\mathcal{E}$\;
	}
	Compute an approximated MCST $\mathcal{G}'$ of $H$\;
	Let $\mathcal{V}_\text{MCMI} \gets \emptyset$, $\mathcal{E}_\text{MCMI} \gets \emptyset$\;
	\For{each edge $e_{(u, v)} \in \mathcal{G}'$}{
		Add nodes and edges from shortest path $P_{(u, v)}$ in MCST to $\mathcal{V}_\text{MCMI}$ and $\mathcal{E}_\text{MCMI}$\;
	}
	Calculate $r(\mathcal{G}_{\text{MCMI}})$ with Equation \eqref{eq:r_mcmi}\;
	\While{the node $v$ in MCMI's 1-hop neighborhood with lowest $c_{e_{(\text{MCMI}, v)}}/ s_v $ satisfies $c_{e_{(\text{MCMI}, v)}}/s_v  < r(\mathcal{G}_{\text{MCMI}})$}{
		Recalculate $r(\mathcal{G}_{\text{MCMI}})$ with Equation \eqref{eq:r_mcmi}\;
		$\mathcal{V}_\text{MCMI} \gets \mathcal{V}_\text{MCMI} \cup \{v\}$\;
		$\mathcal{E}_\text{MCMI} \gets \mathcal{E}_\text{MCMI} \cup \{e_{(\text{MCMI}, v)}\}$\;
	}
	\Return{$\mathcal{G}_\text{MCMI}(\mathcal{V}_\text{MCMI}, \mathcal{E}_\text{MCMI})$}\;
\end{algorithm}
\subsection{Step 3: Hybrid Retrieval}\label{sec:oi}
{% \color{blue}
Other than Graph retrieval, we also apply Hybrid Retrieval (HR) to further enrich the context for query answering. 
In Hybrid Retrieval, we retrieve the most similar text chunks $\mathcal{T}_{HR}$ from all text chunks $t\in\mathcal{T}_c$ to the query's vector representation $\textbf{q}$, based on the hybrid retrieval score $\text{HS}(\cdot)$:
\begin{align}
	&\mathcal{T}_\text{HR} = \left\{ t \in \mathcal{T}_{c}\,\middle|\, t \in \arg\top_{k_r} \left(\text{HS}({q},{t}) \right) \right\},\\ &\text{HS}({q},{t})=\frac{\text{MS}(\textbf{q},\textbf{t})+\text{BM25}(q,t)}{2}
\end{align}
% \footnote{ check the notations}
where $k_r$ denotes the hyperparameter that controls the number of hybrid retrieved text chunks, $\textbf{t}$, $\textbf{q}$ denotes the vector representation of text chunk $t$ and query $q$, respectively; $\text{HS}(q,t)$, $\text{BM25}(q,t)$ and $\text{MS}(\textbf{q},\textbf{t})$ denotes the hybrid retrieval score, the BM25 \cite{robertson2009probabilistic} sparse retrieval score, and the dense representation vectors' cosine similarity score between query $q$ and text $t$ in Equation \eqref{eq:cos_sim}.

After obtaining the textual MCMI subgraph string $g$ and the chunks $\mathcal{T}_\text{HR}$ from hybrid retrieval, 
we input them together with the query $q$ and 
% the text graph string $g$, text chunks $\mathcal{T}_\text{HR}$ from hybrid retrieval (HR), and 
the query answer generation LLM instruction $p_\text{gen}$, to ask the LLM to give an answer to query $q$ based on these provided contents.
\begin{equation}
	\label{eq:qa_prompt}
	\text{answer}=\text{LLM}(q, g, \mathcal{T}_\text{HR},p_\text{gen}), 
\end{equation}

\subsection{Complexity Analysis}
\label{sec:complexity}
\subsubsection{\textbf{Time Complexity}}
% \noindent \textbf{Graph Construction Complexity} 
% Firstly, we analysis the time complexity of AGRAG's Step 1 and Step 2. 
For the entity extraction procedure, the time complexity is $O(|\mathcal{V}||\mathcal{T}_c|)$; 
Adding passage nodes also requires $O(|\mathcal{V}||\mathcal{T}_c|)$ time. 
For graph weighting, calculating the PPR score requires $O(|\mathcal{T}_c|(|\mathcal{V}|+|\mathcal{E}|))$ time, and calculate the edge cost requires $O(|\mathcal{E}|)$ time. 
% Hence, the total time complexity of AGRAG's Graph Indexing and Graph Weghting step would be: 
% $O(|\mathcal{T}_c|(|\mathcal{V}|+|\mathcal{E}|))$. 
For AGRAG's Step 3, 
the time complexity of AGRAG's MCST generation is $O(|\mathcal{E}|+|\mathcal{V}|log|\mathcal{V}|)$ using the Mehlhorn algorithm \cite{mehlhorn1988faster}; 
The expansion from MCST to MCMI would cost 
$O(|\mathcal{V}_\text{MCMI}|\cdot|\mathcal{N}(\mathcal{V}_\text{MCMI})|)$ Hence, the time complexity of AGRAG's MCMI generation at the worst case when $\mathcal{V}_\text{MCMI}=\mathcal{V}$ and the graph is fully connected is:
$O(|\mathcal{V}|^2+|\mathcal{E}|).$

Hence, the {end to end time complexity} of AGRAG is:
	$O(|\mathcal{T}_c|(|\mathcal{V}|+|\mathcal{E}|)+|\mathcal{V}|^2)$. 
% {\color{red}
Since the number of text chunk $|\mathcal{T}_c|$ is less or equal to the number of nodes $|\mathcal{V}|$ in graph, in the worst case,  {the end to end time complexity} of AGRAG is:
\begin{center}
	$O(|\mathcal{V}|\cdot|\mathcal{E}|+|\mathcal{V}|^2)$. 
\end{center}

\subsubsection{\textbf{Space Complexity}}
% For the space complexity of AGRAG's Step 1 and Step 2. 
The graph of AGRAG's Step 1 and Step 2 is stored with the dense weighted adjacency matrix, which needs $O(|\mathcal{E}|)$ space; Storing the text chunk would need $O(|\mathcal{T}_c|)$ space; 
% If the label names are available, 
Storing the entity nodes require $O(|\mathcal{V}|)$ space. Hence the total space complexity of AGRAG's Graph Indexing step is: 
$O(|\mathcal{V}|+|\mathcal{E}|+|\mathcal{T}_c|)$. 
In AGRAG's Step 3, we only need $O(|\mathcal{E}|)$ to store the graph. 
% Hence, the total space complexity of AGRAG's Retrieval \& Generation procedure is 
%$O(|\mathcal{E}|).$

Hence, the total space complexity of AGRAG is:
\begin{center}
	$O(|\mathcal{V}|+|\mathcal{E}|+|\mathcal{T}_c|)$. 
\end{center}

\vspace{-5px}
\section{Experiments}\label{sec:experiment}
	In this section, 
	% we present the experimental evaluation of our framework AGRAG against 6 effective baselines spanning 3 technical categories across 5 tasks. 
	 % We compare the performance of AGRAG against six effective baselines spanning three technical categories. 
	% Specifically, 
	we first outline the experimental setup in Section~\ref{ssec:exp:setting}, including details on datasets, and evaluation metrics. 
	% Due to the space limitation, we put the details of baselines, and hyperparameter configurations in Appendix Section \ref{sec:exp_detail}.
	Next, we report the experimental results, focusing on both effectiveness and efficiency evaluations, in Section~\ref{sec:effectiveness} and Section~\ref{sec:efficiency}, respectively.
	Finally, we conduct ablation study in Section \ref{ssec:exp:ablation}, and parameter sensitivity study in Section \ref{sec:para_sen}.
% 	}
%  \vspace{-15px}
\subsection{Experiment Settings}\label{ssec:exp:setting}
 
\subsubsection{\textbf{Datasets \& Tasks}}
We select the Novel and Medical datasets from GraphRAG-bench \cite{xiang2025use} to evaluate AGRAG's performance on 4 tasks that have different complexity levels:
% these tasks have the same input and output format with the classic QA task:
\begin{itemize}[leftmargin=*]
	\item \textbf{Fact Retrieval}: The simplest (level 1) task of GraphRAG-bench, to solve this task, the model needs to retrieve isolated facts in the graph with minimal reasoning needed. This task mainly tests the models' precise keyword matching ability.
	\item \textbf{Complex Reasoning (Complex Rea.)}: The level 2 task of GraphRAG-bench, to solve this task, the model needs to connect multiple facts on the graph to form multi-hop, long-range reasoning chains.
	\item \textbf{Contextual Summarization (Contextual Sum.)}: The level 3 task of GraphRAG-bench. Here, the model needs to synthesize fragmented information from the graph into comprehensive summaries. 
	% As the summary is not the longer the better, in our experiment, we ask the model to generate the summary within 4096 tokens, and truncate the generated summary to 4096 tokens for evaluation. 
	\item \textbf{Creative Generation (Creative Gen.)}: The most complex (level 4) task of GraphRAG-bench, to solve this task, the model needs to generate hypothetical responses based on but beyond the retrieved knowledge. 
	% Similar to the setting of the Contextual Summarization task, we also ask the model to generate and evaluate with no more than 4096 tokens.
\end{itemize} 
Furthermore, we also apply the text classification task, 
% which is a classic task to evaluate the ability of traditional PLMs such as BERT \cite{devlin2018bert}, and reflects the real-world industrial usage of language models. 
% In this paper, we apply this task 
to test the RAG models' ability to summarize the topic of text passages, and classify them into the correct category within a given label sets. 
% This task are also considered a complex task, as it also require models' summarization ability.
% detect the correlations between texts and labels without any supervision, as well as their ability to strictly follow instructions and predict within given label sets.

\begin{itemize}[leftmargin=*]
\item \textbf{Text Classification}: Given a set of texts $\mathcal{T}$ and a set of label $\mathcal{L}$, the classification task aim to assign each text $t\in \mathcal{T}$ a label $l\in \mathcal{L}$ that best suited to describe its semantic meaning. For text classification task, we select the Web-Of-Science (WOS) dataset \cite{kowsari2017hdltex}, and the IFS-Rel dataset \cite{xia-etal-2021-incremental}. 
\end{itemize} 
We select one text passage per label from the training sets of the WOS and IFS-Rel datasets to construct the corpus for RAG. During retrieval, instead of retrieving the full text chunk, we enable HippoRAG2 and AGRAG to directly retrieve the corresponding label of each text chunk. The retrieved labels are then used to form a candidate label set for the LLM's text prediction. The LLM is subsequently tasked with selecting the most appropriate label from this candidate set. The detailed dataset and task statistics are shown in Table \ref{tab:data_stat}. 

\begin{table*}[t!]
	\centering   
	\renewcommand{\arraystretch}{1.15}
	\caption{Statistics of the Novel and Medical dataset from GraphRAG-bench \cite{xiang2025use} across 4 tasks, and statistics of the WOS \cite{kowsari2017hdltex} and IFS-REL \cite{xia-etal-2021-incremental} dataset for the text classification task. CG, CS, CR, FR and TC denotes the Creative Generation, Contextual Summarization, Complex Reasoning, Fact Retrieval, and Text Classification task, respectively. `-' denotes the respective task is not included by certain dataset.}
	\label{tab:data_stat}
	\setlength\tabcolsep{1.5pt}
	% \vspace{-10px}
{%
		\begin{tabular}{l|c|c|c|c|c|c|c|c|c|c|c|cc} 
			\hline\hline
			\multirow{2}{*}{\textbf{Dataset}} & \multirow{2}{*}{\textbf{Chunks}} & \multirow{2}{*}{\textbf{KG Entities}} & \multirow{2}{*}{\textbf{KG Relations}} & \multirow{2}{*}{\begin{tabular}[c]{@{}c@{}}\textbf{AVG MCST}\\\textbf{Nodes}\end{tabular}} & \multirow{2}{*}{\begin{tabular}[c]{@{}c@{}}\textbf{AVG MCST}\\\textbf{Edges}\end{tabular}} & \multirow{2}{*}{\begin{tabular}[c]{@{}c@{}}\textbf{AVG MCMI}\\\textbf{Nodes}\end{tabular}} & \multirow{2}{*}{\begin{tabular}[c]{@{}c@{}}\textbf{AVG MCMI}\\\textbf{Edges}\end{tabular}} & \textbf{CG.} & \textbf{CS.} & \textbf{CR.} & \textbf{FR.} & \multicolumn{2}{c}{\textbf{TC.}}  \\ 
			\cline{9-14}
			&                                  &                                       &                                        &                                                                                                    &                                                                                                    &                                                                                            &                                                                                            & \multicolumn{4}{c|}{\textbf{\textbf{Questions}}}          & \textbf{Texts} & \textbf{Labels}  \\ 
			\hline\hline
			Novel                             & 5,140                            & 63,887                                & 163,169                                & 8.7                                                                                                & 10.7                                                                                               & 19.7                                                                                       & 35.9                                                                                       & 67           & 362          & 610          & 971          & -              & -                \\
			Medical                           & 1,030                            & 12,426                                & 52,054                                 & 8.9                                                                                                & 9.9                                                                                                & 23.7                                                                                       & 59.5                                                                                       & 166          & 289          & 509          & 1,098        & -              & -                \\
			WOS                               & 24                               & 475                                   & 923                                    & 15.5                                                                                               & 17.8                                                                                               & 23.8                                                                                       & 35.8                                                                                       & -            & -            & -            & -            & 9,396          & 133              \\
			IFS-Rel                           & 11                               & 185                                   & 355                                    & 11.0                                                                                               & 12.8                                                                                               & 17.0                                                                                       & 26.3                                                                                       & -            & -            & -            & -            & 2,560          & 64               \\
			\hline\hline
		\end{tabular}
	}
	% \vspace{-10px}
\end{table*}
\begin{table*}
	\centering
	\renewcommand{\arraystretch}{1.2}
	\caption{Model effectiveness on 4 datasets for 5 tasks, all experiments were carried out with Qwen2.5-14B-Instruct LLM backbone.	}
	\label{tab:main_exp}
	\vspace{-5px}
	\setlength\tabcolsep{1.5pt}
	\resizebox{\linewidth}{!}{%
\begin{tabular}{l|l|ccc|ccc|cc|cc|cc|cc|cc|cc|cc|cc} 
	\hline\hline
	\multirow{3}{*}{\textbf{Category }}                                            & \textbf{Task}    & \multicolumn{6}{c|}{\textbf{Creative Gen. }}                                                        & \multicolumn{4}{c|}{\textbf{Contextual Sum. }}                                & \multicolumn{4}{c|}{\textbf{Complex Rea. }}                                   & \multicolumn{4}{c|}{\textbf{Fact Retrieval }}                                 & \multicolumn{4}{c}{\textbf{Text Classification }}                           \\ 
	\cline{2-24}
	& \textbf{Dataset} & \multicolumn{3}{c|}{\textbf{Novel }}             & \multicolumn{3}{c|}{\textbf{Medical}}            & \multicolumn{2}{c|}{\textbf{Novel }} & \multicolumn{2}{c|}{\textbf{Medical }} & \multicolumn{2}{c|}{\textbf{Novel }} & \multicolumn{2}{c|}{\textbf{Medical }} & \multicolumn{2}{c|}{\textbf{Novel }} & \multicolumn{2}{c|}{\textbf{Medical }} & \multicolumn{2}{c|}{\textbf{WOS }} & \multicolumn{2}{c}{\textbf{IFS-REL }}  \\ 
	\cline{2-24}
	& \textbf{Metric}  & \textbf{FS.}   & \textbf{ACC.}  & \textbf{COV.}  & \textbf{FS.}   & \textbf{ACC.}  & \textbf{COV.}  & \textbf{COV.}  & \textbf{ACC.}       & \textbf{COV.}  & \textbf{ACC.}         & \textbf{ROG.}  & \textbf{ACC.}       & \textbf{ROG.}  & \textbf{ACC.}         & \textbf{ROG.}  & \textbf{ACC.}       & \textbf{ROG.}  & \textbf{ACC.}         & \textbf{ACC.}  & \textbf{REC.}     & \textbf{ACC.}  & \textbf{REC.}         \\ 
	\hline\hline
	\multirow{2}{*}{\begin{tabular}[c]{@{}l@{}}Traditional\\RAG\end{tabular}}      & NaiveRAG         & 0.205          & 0.307          & \uline{0.384}  & 0.176          & \uline{0.577}  & \uline{0.554}  & 0.745          & 0.621               & \uline{0.677}  & 0.576                 & 0.120          & 0.447               & 0.117          & 0.635                 & 0.117          & 0.431               & 0.182          & 0.604                 & 0.237          & 0.257             & 0.351          & 0.197                 \\
	& L.LLMLingua      & 0.213          & 0.316          & 0.377          & 0.225          & 0.575          & 0.496          & 0.687          & 0.627               & 0.657          & 0.583                 & 0.092          & 0.434               & 0.080          & 0.633                 & 0.115          & 0.420               & 0.105          & 0.591                 & 0.274          & 0.231             & 0.378          & 0.220                 \\ 
	\hline
	\multirow{2}{*}{\begin{tabular}[c]{@{}l@{}}Tree\\-based RAG\end{tabular}}      & RAPTOR           & 0.365          & \uline{0.333}  & 0.373          & 0.314          & 0.526          & 0.553          & \uline{0.760}  & 0.626               & 0.633          & 0.583                 & 0.290          & \uline{0.460}       & 0.185          & \uline{0.649}         & 0.133          & 0.465               & 0.132          & 0.609                 & 0.409          & 0.290             & 0.293          & 0.212                 \\
	& SiReRAG          & 0.324          & 0.328          & 0.365          & 0.295          & 0.544          & 0.456          & 0.754          & 0.617               & 0.592          & 0.560                 & 0.307          & 0.459               & 0.269          & 0.646                 & 0.311          & 0.451               & 0.349          & 0.558                 & \uline{0.430}  & \uline{0.294}     & 0.325          & 0.220                 \\ 
	\hline
	\multirow{4}{*}{\begin{tabular}[c]{@{}l@{}}Community\\-based RAG\end{tabular}} & G-retriever      & 0.294          & 0.167          & 0.200          & 0.169          & 0.371          & 0.468          & 0.594          & 0.404               & 0.302          & 0.395                 & 0.146          & 0.441               & 0.102          & 0.396                 & 0.199          & 0.209               & 0.148          & 0.326                 & 0.188          & 0.129             & 0.311          & 0.185                 \\
	& MIXRAG           & 0.200          & 0.286          & 0.337          & 0.255          & 0.351          & 0.306          & 0.695          & 0.538               & 0.409          & 0.464                 & 0.109          & 0.386               & 0.156          & 0.434                 & 0.102          & 0.366               & 0.160          & 0.463                 & 0.244          & 0.235             & 0.362          & 0.203                 \\
	& GraphRAG         & 0.427          & 0.310          & 0.339          & 0.345          & 0.475          & 0.325          & 0.758          & 0.620               & 0.625          & 0.506                 & 0.246          & 0.449               & 0.250          & 0.606                 & 0.240          & 0.445               & 0.217          & 0.521                 & 0.253          & 0.201             & 0.279          & 0.202                 \\
	& LightRAG         & 0.201          & 0.314          & 0.329          & 0.189          & 0.567          & 0.549          & 0.753          & \uline{0.628}       & 0.625          & 0.585                 & 0.223          & 0.448               & 0.255          & 0.637                 & 0.253          & 0.460               & 0.283          & 0.607                 & 0.273          & 0.224             & 0.265          & 0.220                 \\ 
	\hline
	\multirow{3}{*}{\begin{tabular}[c]{@{}l@{}}Walking\\-based RAG\end{tabular}}   & HippoRAG2        & \uline{0.496}  & 0.295          & 0.325          & \uline{0.347}  & 0.563          & 0.451          & 0.735          & 0.586               & 0.496          & 0.560                 & \uline{0.314}  & 0.453               & \uline{0.332}  & 0.622                 & 0.310          & \uline{0.470}       & 0.343          & 0.598                 & 0.424          & 0.270             & \uline{0.389}  & \uline{0.246}         \\
	& F.GraphRAG       & 0.286          & 0.287          & 0.376          & 0.217          & 0.500          & 0.396          & 0.706          & 0.528               & 0.507          & \uline{0.587}         & 0.237          & 0.371               & 0.262          & 0.643                 & \uline{0.318}  & 0.439               & \uline{0.352}  & \uline{0.610}         & 0.283          & 0.245             & 0.307          & 0.221                 \\
	& \textbf{AGRAG}   & \textbf{0.513} & \textbf{0.339} & \textbf{0.386} & \textbf{0.356} & \textbf{0.598} & \textbf{0.589} & \textbf{0.778} & \textbf{0.643}      & \textbf{0.690} & \textbf{0.604}        & \textbf{0.319} & \textbf{0.467}      & \textbf{0.343} & \textbf{0.665}        & \textbf{0.324} & \textbf{0.490}      & \textbf{0.357} & \textbf{0.628}        & \textbf{0.438} & \textbf{0.307}    & \textbf{0.403} & \textbf{0.251}        \\
	\hline\hline
\end{tabular}
	}
	\vspace{-10px}
\end{table*}
\subsubsection{\textbf{Evaluation Metrics}}
\label{ssec:exp:metric}
% {\color{blue}
	In this paper, we apply 5 metrics with the RAGAS \cite{ragas2024} package to evaluate AGRAG's performance on 5 different tasks:
	%}
\begin{enumerate}[leftmargin=*, label=\alph*.]
\item \textbf{Accuracy (ACC.)}: For text classification, the proportion of exact matches between LLM and gold labels. For other tasks,  sample is scored 0, 0.5, or 1 by the LLM for inaccurate, partially accurate, or exactly accurate responses.

%	\item \textbf{Accuracy (ACC.)}: Measures the proportion of correct predictions among all predictions made by a model,  for the text classification task. We calculate this metric based on an exact match between the LLM-generated label and the gold label. For other tasks, we assign each data sample an accuracy score of 0, 0.5, or 1 if the LLM's response is inaccurate, partially accurate, or exactly accurate, with the LLM for score assigning.
	
	\item \textbf{Recall (REC.)}: Quantifies the model's ability to identify all actual positive instances by calculating the ratio of true positives to the sum of true positives and false negatives, we calculate the macro recall score and apply it to the text classification task, where we have a fixed label set.
	\item \textbf{ROUGE-L (ROG.)} \cite{lin2004rouge}: Evaluates the answer quality by comparing the longest common subsequence (LCS) between the generated answer $G$ and the gold answer $Y$: $ \text{ROUGE-L}=\frac{\text{LCS}(G,Y)}{|G|}$.
	 % \begin{equation}
	% 	\text{ROUGE-L}=\frac{\text{LCS}(G,Y)}{|G|}, 
	% \end{equation}
	We apply ROUGE-L to the fact retrieval and the complex reasoning tasks, as the answer of these two tasks are usually a fixed entity or relation.
	\item \textbf{Coverage (COV.)} \cite{xiang2025use}: Evaluates answer completeness by assessing whether all necessary reference evidences $E$ are included in the LLM's response $R$, computed as the proportion of covered evidences to the number of required evidences $|E|$:$\text{Cov}=\frac{|\{e\in E\ |\ M(e,R)\}|}{|E|},$ 
	where $M(e,R)$ is a boolean function indicating whether evidence $e$ is covered by generated response $R$. We apply the coverage metric to contextual summarization and creative generation tasks, as their answers are usually passages that conclude multiple information points.
	
	\item \textbf{Faithfulness (FS.)} \cite{xiang2025use}: Measures the percentage of answer claims $A$ directly supported by retrieved context $C$, calculated as the ratio of claims aligned with evidence to the number of total claims $A$: $\text{FS}=\frac{|\{a\in A\ |\ S(a,C)\}|}{|A|},$
	where $S(a,C)$ is a boolean function indicating whether a claim is supported by contexts in $C$. We apply the faithfulness metric to the creative generation task to evaluate the quality of the hypothesis in the LLM's response.
\end{enumerate}

\begin{table*}[t]
	\centering
	\renewcommand{\arraystretch}{1.2}
	\caption{Model efficiency on 4 datasets for 5 tasks, all experiments were carried out with Qwen2.5-14B-Instruct LLM backbone. 
		%Where AGRAG achieve at most 166.3\% accleration, and 368.8\% less token cost compared with tested baselines.
	}
	\label{tab:efficiency}
	% \vspace{-10px}
	\setlength\tabcolsep{1pt}
 
		\begin{tabular}{l|l|cccc|cccc} 
			\hline\hline
			\multirow{2}{*}{\textbf{Category}}                                             & \multirow{2}{*}{\textbf{Models}} & \multicolumn{4}{c|}{\textbf{Time Cost (Indexing / Querying mins) }}           & \multicolumn{4}{c}{\textbf{Token Cost (Input / Output per query) }}                                             \\ 
			\cline{3-10}
			&                                  & \textbf{Novel}    & \textbf{Medical}  & \textbf{WOS}      & \textbf{IFS-REL}  & \textbf{Novel}               & \textbf{Medical}             & \textbf{WOS}               & \textbf{IFS-REL}     \\ 
			\hline\hline
			\multirow{2}{*}{\begin{tabular}[c]{@{}l@{}}Traditional\\RAG\end{tabular}}      & NaiveRAG                         & \textbf{13} / 371 & \textbf{7} / 245  & \textbf{2} / 292  & \textbf{2} / 225  & 9,437 / 1,708                & 7,473 / 379                  & 7,292 / 29                  & 4,521 / \uline{16}   \\
			& L.LLMLingua                      & 51 / 360          & 37 / 305          & \uline{4} / 358   & \uline{3} / 315   & 8,466 / 1,750                & 5,686 / 390                  & 5,861 / \uline{31}         & 2,757 / \textbf{14}  \\ 
			\hline
			\multirow{2}{*}{\begin{tabular}[c]{@{}l@{}}Tree\\-based RAG\end{tabular}}      & RAPTOR                           & 464 / 250         & 117 / 319         & 30 / 162          & 24 / \textbf{121} & 7,243 / 777                  & \uline{4,817} / 695          & 6,007 / 41                 & 3,807 / 33           \\
			& SiReRAG                          & 85 / 233          & 40 / 207          & 26 / 139          & 18 / 117          & 7,759 / 800                  & 4,918 / 701                  & 6,012 / 47                 & 3,821 / 36           \\ 
			\hline
			\multirow{4}{*}{\begin{tabular}[c]{@{}l@{}}Community\\-based RAG\end{tabular}} & G-Retriever                      & 19 / \textbf{119} & 12 / 145          & 8 / \uline{158}   & \uline{3} / 164   & \uline{6,344} / \textbf{116} & 6,290 / \textbf{157}         & 6,762 / \textbf{12}        & 4,002 / 21           \\
			& MIXRAG                           & \uline{17} / 388  & \uline{11} / 453  & 5 /  257          & \textbf{2} / 193  & \textbf{6,083} / \uline{237} & \textbf{4,238 }/ \uline{230} & \uline{4,193 }/ \uline{31} & 3,597 / 29           \\
			& GraphRAG                         & 347 / 205         & 194 / 158         & 85 / 240          & 64 / 217          & 12,686 / 2,109               & 8,385 / 1,017                & 11,809 / 351               & 9,009 / 287          \\
			& LightRAG                         & 166 / 221         & 116 / 164         & 42 / 211          & 31 / 197          & 13,145 / 1,233               & 9,535 / 426                  & 11,510 / 227               & 7,667 / 199          \\ 
			\hline
			\multirow{3}{*}{\begin{tabular}[c]{@{}l@{}}Walking\\-based RAG~\end{tabular}}  & HippoRAG2                        & 95 / \uline{134}  & 46 / \textbf{130} & 19 / 140          & 15 / 127          & 8,138 / 1,073                & 6,264 / 595                  & 5,066 / 93                 & 2,440 / 102          \\
			& F.GraphRAG                       & 106 / 198         & 48 / 173          & 29 / 186          & 27 / 164          & 7,490 / 1,252                & 6,475 / 384                  & 4,957 / 35                 & \uline{2,217} / 20   \\
			& \textbf{AGRAG}                   & 65 / 142          & 16 / \uline{138}  & 16 / \textbf{132} & 14 / \uline{124}  & 6,487 / 1,140                & 5,145 / 413                  & \textbf{3,970} / 101       & \textbf{1,890} / 93  \\
			\hline\hline
		\end{tabular}
		\vspace{-15px}
	
\end{table*}
\subsubsection{\textbf{Baselines}}
% As shown in Table \ref{tab:main_exp}, 
We compare AGRAG's performance with 6 baselines from 3 technical categories, they are traditional RAG model NaiveRAG \cite{gao2023retrieval} and LongLLMLingua \cite{jiang-etal-2024-longllmlingua}; Community-based model GraphRAG \cite{edge2024local}, and LightRAG \cite{guo2024lightrag}; And walking-based model Fast-GraphRAG \cite{fastgraphrag} and HippoRAG2 \cite{gutiérrez2025ragmemory}.
{% \color{blue}
We use the Qwen-2.5-14B-Instruct \cite{qwen2025qwen25technicalreport} as the LLM backbone unless further illustrated.
}

\noindent \textbf{Traditional RAG Models}
These models directly retrieve text chunks based on their semantic similarity to the queries.
\begin{itemize}[leftmargin=10pt]
	\item  \textbf{NaiveRAG} \cite{gao2023retrieval}: NaiveRAG acts as a foundational baseline of current RAG models. 
	It encodes text chunks into a vector database and retrieve the Top-k text chunks based on the vector similarities between queries and text chunks.
	% When indexing, it stores text chunks segmented from the text corpus in a vector database using text embeddings. When retrieving, NaiveRAG generates query texts' vectorized representations to retrieve a fixed number of text chunks based on the embedding similarities.
	\item \textbf{L.LLMLingua} \cite{jiang-etal-2024-longllmlingua}: 
	% LongLLMLingua is an instruction-aware prompt compressor model. 
	LongLLMLingua utilizes LLM's generation perplexity to filter out unimportant LLM input tokens. 
	% based on the retrieved side information and the task instruction. 
	In this paper, we apply it to compress the retrieved text chunks of NaiveRAG w.r.t. the query and task instruction.
	% to test the performance of a simple non-Graph-based RAG model when applying a straightforward prompt compressor.
\end{itemize} 
% {\color{blue}
\noindent \textbf{Tree-based Models} 
These models organize text chunks into tree structures to capture long-range dependencies and thematic information.
\begin{itemize}[leftmargin=10pt]
	\item \textbf{RAPTOR} \cite{sarthi2024raptor}
	RAPTOR recursively clusters and summarizes text chunks to construct a tree from the bottom up, enabling the retrieval of information at varying levels of abstraction to answer high-level thematic queries.
	
	\item \textbf{SiReRAG} \cite{zhang2025sirerag}
Based on RAPTOR, SiReRAG builds similarity and relatedness trees to index respective information, thereby supporting complex multihop reasoning.

\end{itemize} 
% }

\noindent \textbf{Community-based Models}
These models retrieve one or multi-hop neighborhood of certain graph nodes that correspond to the query.
\begin{itemize}[leftmargin=10pt]
	\item \textbf{G-retriever} \cite{he2024g} 
		It formulates subgraph retrieval as a Prize-Collecting Steiner Tree (PCST) optimization problem, identifying connected subgraphs that maximize relevance while minimizing costs.
	
	\item \textbf{MIXRAG} \cite{liu2025mixrag}
		It introduces a Mixture-of-Experts (MoE) framework to handle diverse query intents, dynamically routing queries to specialized experts and employing a query-aware graph encoder to filter noise.

	\item \textbf{GraphRAG} \cite{edge2024local}: It employs LLM for graph construction. When retrieval, it aggregates nodes into communities w.r.t. the query, and generates community reports to encapsulate global information from texts.
	
	\item \textbf{LightRAG} \cite{guo2024lightrag}: 
	% It skips the GraphRAG's explicit formulation of graph communities and the community summary. 
	It calculates the vector similarity between query extracted entities and graph nodes, achieving a one-to-one mapping from entities to graph nodes, then retrieve these nodes' graph neighborhood.
\end{itemize} 
\noindent \textbf{Walking-based Models}
These models apply walking-based graph algorithms to score graph nodes, then retrieve text chunks w.r.t. their contained entity nodes' score.
\begin{itemize}[leftmargin=10pt]
\item \textbf{F.GraphRAG} \cite{fastgraphrag}: 
%Fast-GraphRAG is a lightweight version of GraphRAG; 
Fast-Graphrag employs LLM for graph construction.
% but 
% It substitutes GraphRAG's community-related operations with the PPR graph traverse mechanism. 
During retrieval, it applies PPR to assign node scores based on the query, 
% traverse graphs and perform multi-hop reasoning over the KG by assigning node scores, 
and retrieve the Top-k passages based on the sum of their contained nodes' scores.
% then it considers the score sum of the passage contained text nodes as the retrieval score of passages, and finally retrieves the passages with the top-k highest retrieval score. 

\item \textbf{HippoRAG2} \cite{gutiérrez2025ragmemory}: 
% HippoRAG2 is a lightweight walking-based model. 
It employs LLM for graph construction, and also add the text chunks as nodes in the graph.  
% It first index corpus extracted triples and the text chunks to form a KG. 
As a result, it directly retrieves Top-k text chunks based on their respective nodes' PPR score.

\end{itemize} 

% Further illustrations of these chosen baselines are listed in \autoref{sec:baselines} in Appendix. 
%For GraphRAG and NaiveRAG, we use the implementation from \cite{nanographrag} which optimizes their original code and achieves better time efficiency while not affect the performance; 
% For BERT PLM, we use the implementation from Hugging Face \cite{wolf2020transformers}; 
%For all other baselines, we use their official implementations. 
% {\color{blue}

%\begin{figure*}[ht]
%	\includegraphics[width=1\linewidth]{figures/graph_statistic.png}
%	\vspace{-20px}
%	\caption{The per corpus graph size of tested Graph-based RAG models on 4 datasets.}
%	\label{fig:graph_size}
% \vspace{-10px}
%\end{figure*}

\subsubsection{\textbf{Hyperparameter and Hardware Settings}}
We set the retrieval top-k value for all RAG models to 5, aligning with the settings of Fast-GraphRAG and HippoRAG2. 
% For LongLLMLingua, we test the compression rate within ${0.75, 0.8, 0.85}$ and select 0.8, as it achieves the best overall classification accuracy.
For GraphRAG, we apply its local search mode to all tasks, as it achieves the overall best effectiveness. 
For LightRAG, we apply its hybrid search mode for all tasks except the text classification task, as it is claimed as one of the contributions of LightRAG. 
% We apply LightRAG's local search mode for the text classification task, as it achieves the best effectiveness on the two datasets tested. 
For Fast-GraphRAG and HippoRAG2, we set their hyperparameters to default, to align with their original paper and code. 
Other hyperparameters of models were set as default by their codes.

{% \color{blue}
For our AGRAG, the length of text chunks $l_t$ is set to 256, and token overlap between chunks $l_o$ to 32, the number of raw mapped triples $k_a$ and the number of hybrid retrieved chunks $k_r$ are both set to 5, to align with the setting of HippoRAG2 and GraphRAG-bench. 
AGRAG's unique hyperparameter, the entity extraction threshold, and the maximum n-gram, were set to 0.5 and 3, respectively. 
Please refer to Section \ref{sec:para_sen} for further parameter sensitivity analysis.
%\footnote{\# notations are wrong Yubo: Fixed}
}

All our experiments were carried out and evaluated with Qwen2.5-14B-Instruct \cite{qwen2025qwen25technicalreport} LLM backbone deployed with vLLM \cite{kwon2023efficient} if not further illustrated. 
We apply the Contriever \cite{izacardunsupervised} embedding model for HippoRAG2 and AGRAG's dense passage retrieval procedure, and bge-large-en-v1.5 \cite{bge_embedding} embedding model for other baselines' LLM embedding. 
We apply Qwen-3-Reranker-0.6B LLM for raw triplet filtering.
All experiments are conducted on an Intel(R) Xeon(R) Gold 5220R CPU and a single NVIDIA A100-80GB GPU.

\subsection{Effectiveness Evaluation}\label{sec:effectiveness}
%\footnote{\# put them into the corresponding experiment part Yubo: Fixed, put it here}
\begin{comment}

	Other than Qwen-2.5-14B-Instruct, we also test the performance of AGRAG with GPT-4o \cite{hurst2024gpt} and the reinforcement learning based RAG reasoning backbone R1-Searcher-Qwen-2.5-7B \cite{song2025r1} in Table \ref{tab:gpt_exp}. 
	Due to cost and space considerations, 
	we only compare AGRAG's performance with the previous state-of-the-art model HippoRAG2 \cite{gutiérrez2025ragmemory} on \textcolor{blue}{Novel and Medical datasets from GraphRAG-bench \cite{xiang2025use}} for GPT-4o, as the text classification task has a large number of texts to be classified, and can lead to high token cost (Over 7,000 USD on WOS dataset, \$15 for 1 million tokens). 
	For R1-Searcher, we further compare NaiveRAG \cite{gao2023retrieval}, as it is the RAG framework used in its original paper \cite{song2025r1}.
\end{comment}
Based on the experiment results in Table \ref{tab:main_exp} 
% and Table \ref{tab:gpt_exp}, 
% As shown in Table \ref{tab:main_exp} and Table \ref{tab:gpt_exp}, 
% AGRAG achieves the overall best performance over all 5 tasks spanning through 4 datasets surpasses all test baselines. 
% Compared with the previous state-of-the art models, it achieves at most 13.6\% gain on text classification task, on other tasks, AGRAG can still achieve at least 3\% performance gain than the previous state-of-the art models. 
% Based on the experiment results, 
we have observations as follow: 
% \vspace{-3px}
% \begin{itemize}[leftmargin=*]

Firstly, in line with recent findings \cite{xiang2025use,han2025rag,zhou2025depth}, NaiveRAG \cite{gao2023retrieval} 
% still outperforms some advanced models on certain tasks. In our experiments, it 
achieves better performance than all tested walking-based baselines on Contextual Summarization and Creative Generation tasks.
% These walking-based baselines utilize the graph as a scoring mechanism, leveraging fine-grained phrase-level information and the structural properties of the graph to score and retrieve text chunks. 
This is because, walking-based baselines overlook the global, and coarse-grained information that is essential for effective summarization and creative generation. 
% Their suboptimal performance in Contextual Summarization 
% This result also suggests that the fine-grained, phrase-level information they capture is less beneficial for summarization compared to higher-level, coarse-grained insights, which help the LLM form a holistic understanding of the text chunks.
% In comparison, Community-based models achieve performance comparable to NaiveRAG on Contextual Summarization tasks. Their graph-generated summaries also support the LLM's overall understanding of the text. However, they may still propagate errors due to inaccuracies in the summary generation process. 
AGRAG achieves the best performance on these tasks by providing the reasoning subgraph, as well as the hybrid retrieved text chunks to the LLM, offering more comprehensive coarse to fine-grained context for query answering.

Secondly, 
% on simpler Complex Reasoning and Fact Retrieval tasks, the more recent walking-based models achieve the best performance.
% This indicates that the phrase-level dependencies, both long and short range, captured by these models' graph scoring mechanisms, are beneficial for solving tasks that require forming a reasoning chain from one or multiple facts to provide factual information. 
% As such chains can be obscured by numerous noisy phrases within a text chunk due to information compression happened in text chunk encoding, the graph scoring mechanism employed by walking-based models outperforms the similarity based retrieval used in NaiveRAG, as they directly score passages based on their structural information within the graph. 
% In the case of Community-based models, their retrieval process can be viewed as generating a summary of a specific graph community with respect to the query. This approach performs better than NaiveRAG on Fact Retrieval and Complex Reasoning tasks, as the summary can focus more effectively on the critical chain of information.
 % However, the relevant chain information may still be overwhelmed by noise within the community. Moreover, any noise introduced during the LLM-based summary generation can propagate and negatively impact later answer generation. 
Our AGRAG achieves the best performance on Fact Retrieval and Complex Reasoning tasks. 
% Unlike other walking-based models that rely solely on inputting text chunks, 
as it can benefit from phrase-level dependencies by generating explicit reasoning chains and directly feeding it to the LLM as guidance.

\begin{table}[t]
	\centering
	\caption{The hallucinated prediction rate on WOS dataset.}
	\label{tab:exp_hal}
	\vspace{-5px}
	{%
		\begin{tabular}{l|l|c} 
			\hline\hline
			\textbf{Category }                                                             & \textbf{Model }         & \textbf{Hallucinated Rate}       \\ 
			\hline
			\multirow{1}{*}{Pure LLM}      & Qwen-2.5-14B                &          36.88\%        \\ \hline
			\multirow{2}{*}{Traditional RAG}      & NaiveRAG                & 9.96\%                    \\
			& L.LLMLingua             & 21.96\%                   \\ 
			\hline
			\multirow{2}{*}{Tree-based RAG}       & RAPTOR                  & 7.40\%                    \\
			& SiReRAG                 & 7.36\%                    \\ 
			\hline
			\multirow{2}{*}{Community-based RAG} & GraphRAG                & 6.90\%                    \\
			& LightRAG                & 5.89\%                    \\ 
			\hline
			\multirow{3}{*}{Walking-based RAG}                                               & HippoRAG2               &                4.25\%           \\
			& F.GraphRAG              &          6.06\%                 \\
			& {\textbf{AGRAG}} & {\textbf{0.40\%}}  \\
			\hline\hline
		\end{tabular}
	}
	\vspace{-10px}
\end{table}

% Thirdly, for the Creative Generation task, NaiveRAG \cite{gao2023retrieval} and HippoRAG2 \cite{gutiérrez2025ragmemory} achieve the best performance among all tested baselines.
 %Similar to the Contextual Summarization task, NaiveRAG attains the highest coverage score among all models except AGRAG, which we attribute to its reliance on global, coarse-grained information from the original text chunks.
 %On the other hand, consistent with their performance on Fact Retrieval and Complex Reasoning tasks, both HippoRAG and HippoRAG2 achieve the highest faithfulness scores again outperforming all models except AGRAG, by effectively capturing phrase-level dependencies.
% AGRAG achieves the best overall performance on the Creative Generation task by combining the strengths of both NaiveRAG and HippoRAG2. It not only incorporates text chunks but also captures phrase-level dependencies through the generation of a reasoning subgraph.

Lastly, in the text classification task, HippoRAG2 \cite{gutiérrez2025ragmemory} and SiReRAG \cite{zhang2025sirerag} achieve the second-best performance across all metrics on WOS and IFS-Rel dataset respectively, outperformed by AGRAG.
This is because text classification requires capturing correlations between specific text entities and the text chunks to be classified, and all of SiReRAG, HippoRAG2 and AGRAG explicitly model entity-to-passage relationships in the KG. 
% As shown in Figure \ref{fig:graph_size}, HippoRAG2 and AGRAG construct a more fine-grained KG with significantly more components than other Graph-based RAG models. 
AGRAG achieves the best performance as 
% it generates a reasoning subgraph and feeds it to the LLM to assist in classification. 
its reasoning subgraph spans multiple entity nodes in the KG and explicitly captures the relationships between these entities and the text chunks associated with a given classification label, which can assist LLM classification.

Furthermore, Table \ref{tab:exp_hal} presents the hallucinated prediction rate, defined as the percentage of classifications out of the label set. AGRAG achieves the lowest rate by leveraging MCMI subgraphs to provide explicit evidence for retrieved chunks.
% \end{itemize}
% \vspace{-10px}$$\text{Table \ref{tab:exp_hal_main} presents the hallucinated prediction rate, defined as the percentage of classifications outside the label set.}

\subsection{Efficiency Evaluation}\label{sec:efficiency}

\begin{table*}
	\centering
	\caption{The ablation experiment result on the two datasets and 4 tasks. All experiments were carried out with Qwen2.5-14B-Instruct LLM, where Time (I / Q) denotes the Index and Query time, Token (I / O) denotes the LLM Input and Output token count.}
	\label{tab:abl_exp}
	\setlength\tabcolsep{1.5pt}
	\resizebox{\linewidth}{!}{%
		\begin{tabular}{l|ccc|ccc|cc|cc|cc|cc|cc|cc|cc|cc} 
			\hline\hline
			\multirow{3}{*}{\textbf{Model }} & \multicolumn{6}{c|}{\textbf{Creative Gen. }}                                                        & \multicolumn{4}{c|}{\textbf{Contextual Sum. }}                              & \multicolumn{4}{c|}{\textbf{Complex Rea. }}                                 & \multicolumn{4}{c|}{\textbf{Fact Retrieval }}                               & \multicolumn{4}{c}{\textbf{Model Efficiency }}                                                                             \\ 
			\cline{2-23}
			& \multicolumn{3}{c|}{\textbf{Novel }}             & \multicolumn{3}{c|}{\textbf{Medical }}           & \multicolumn{2}{c|}{\textbf{Novel}} & \multicolumn{2}{c|}{\textbf{Medical}} & \multicolumn{2}{c|}{\textbf{Novel}} & \multicolumn{2}{c|}{\textbf{Medical}} & \multicolumn{2}{c|}{\textbf{Novel}} & \multicolumn{2}{c|}{\textbf{Medical}} & \multicolumn{2}{c|}{\textbf{Novel }}                         & \multicolumn{2}{c}{\textbf{Medical }}                       \\ 
			\cline{2-23}
			& \textbf{FS.}   & \textbf{ACC.}  & \textbf{COV.}  & \textbf{FS.}   & \textbf{ACC.}  & \textbf{COV.}  & \textbf{COV. } & \textbf{ACC.}      & \textbf{COV.}  & \textbf{ACC.}        & \textbf{ROG.}  & \textbf{ACC.}      & \textbf{ROG.}  & \textbf{ACC.}        & \textbf{ROG.}  & \textbf{ACC.}      & \textbf{ROG.}  & \textbf{ACC.}        & \textbf{Time (I / Q)}         & \textbf{Token (I / G)}       & \textbf{Time (I / Q)}        & \textbf{Token (I / G)}       \\ 
			\hline\hline
			AGRAG w. HR Only                 & 0.491          & \uline{0.322}  & 0.378          & 0.303          & 0.526          & 0.549          & 0.739          & 0.522              & 0.535          & 0.583                & 0.303          & 0.453              & 0.334          & \uline{0.636}        & \uline{0.323}  & \uline{0.487}      & 0.355          & 0.619                & \textbf{11.7} / \uline{137.9} & \textbf{5,117} / \uline{193} & 11.4 / \uline{117.9}         & \uline{5,215} / \uline{199}  \\
			AGRAG w. LLM ER                  & 0.428          & 0.302          & \uline{0.365}  & 0.352          & 0.526          & \uline{0.587}  & \uline{0.755}  & 0.552              & 0.600          & 0.566                & \uline{0.320}  & \uline{0.465}      & 0.333          & 0.624                & 0.322          & 0.480              & 0.353          & 0.613                & 83.3 /  161.2                 & 8,030 /   887                & 31.9 / 148.8                 & 5,457  / 414                 \\
			AGRAG w.o. HR                    & \textbf{0.519} & 0.306          & 0.343          & 0.342          & 0.521          & 0.570          & 0.677          & 0.534              & 0.420          & 0.578                & 0.318          & 0.434              & 0.324          & 0.598                & 0.314          & \uline{0.487}      & 0.344          & 0.615                & \uline{61.7} / 138.9          & 6,986 / 741                  & \uline{9.4 }/ \textbf{114.4} & \textbf{4,820} / 315         \\
			AGRAG w.o. MCMI                  & 0.488          & \uline{0.332}  & 0.358          & \uline{0.351}  & \uline{0.530}  & 0.575          & 0.748          & \uline{0.592}      & 0.600          & \uline{0.588}        & \textbf{0.324} & \uline{0.465}      & 0.329          & 0.633                & 0.322          & 0.470              & \textbf{0.359} & \uline{0.621}        & 63.9 / \textbf{130.4}         & \uline{6,710} / 757          & 9.5 /  125.0                 & 5,230 /    331               \\
			AGRAG w.o. Cle.                  & 0.478          & 0.311          & 0.361          & 0.336          & 0.517          & 0.569          & 0.712          & 0.597              & \uline{0.616}  & 0.577                & 0.301          & 0.450              & \uline{0.334}  & 0.628                & 0.313          & 0.478              & 0.354          & 0.620                & 65.6 / 140.1                  & 6,988 / \textbf{154}         & 9.1 / 142.9                  & 5,252 / \textbf{158}         \\
			% AGRAG-L                          & 0.441          & 0.519          & 0.354          & 0.336          & 0.777          & 0.490          & 0.745          & 0.706              & 0.545          & 0.751                & 0.293          & 0.552              & 0.312          & 0.739                & 0.297          & 0.560              & 0.351          & 0.701                & 44.7 / 132.1                  & 6,625 / 1,172                & 14.4 / 167.3                 & 5,282 / 534                  \\
			\textbf{AGRAG}                   & \uline{0.513}  & \textbf{0.339} & \textbf{0.386} & \textbf{0.356} & \textbf{0.598} & \textbf{0.589} & \textbf{0.778} & \textbf{0.643}     & \textbf{0.690} & \textbf{0.604}       & 0.319          & \textbf{0.467}     & \textbf{0.343} & \textbf{0.665}       & \textbf{0.324} & \textbf{0.490}     & \uline{0.357}  & \textbf{0.628}       & 65.6 / 142.0                  & 6,988 / 639                  & \textbf{9.1} / 145.3         & 5,252 / 306                  \\
			\hline\hline
		\end{tabular}
	}
	\vspace{-5px}
\end{table*}

\begin{table*}
	\centering
	\caption{Low resource experiment effectiveness with Qwen3-4B-Instruct-2507 LLM on 2080Ti GPU, where model with -L denotes low resource setting.}
	\label{tab:exp_low}
	\vspace{-5px}
	\setlength\tabcolsep{1.5pt}{%
		\begin{tabular}{l|ccc|ccc|cc|cc|cc|cc|cc|cc} 
			\hline\hline
			\textbf{Task}    & \multicolumn{6}{c|}{\textbf{Creative Gen. }}                                                        & \multicolumn{4}{c|}{\textbf{Contextual Sum. }}                                & \multicolumn{4}{c|}{\textbf{Complex Rea. }}                                   & \multicolumn{4}{c}{\textbf{Fact Retrieval }}                                  \\ 
			\hline
			\textbf{Dataset} & \multicolumn{3}{c|}{\textbf{Novel }}             & \multicolumn{3}{c|}{\textbf{Medical }}           & \multicolumn{2}{c|}{\textbf{Novel }} & \multicolumn{2}{c|}{\textbf{Medical }} & \multicolumn{2}{c|}{\textbf{Novel }} & \multicolumn{2}{c|}{\textbf{Medical }} & \multicolumn{2}{c|}{\textbf{Novel }} & \multicolumn{2}{c}{\textbf{Medical }}  \\ 
			\hline
			\textbf{Metric}  & \textbf{FS.}   & \textbf{ACC.}  & \textbf{COV.}  & \textbf{FS.}   & \textbf{ACC.}  & \textbf{COV.}  & \textbf{COV.}  & \textbf{ACC.}       & \textbf{COV.}  & \textbf{ACC.}         & \textbf{ROG.}  & \textbf{ACC.}       & \textbf{ROG.}  & \textbf{ACC.}         & \textbf{ROG.}  & \textbf{ACC.}       & \textbf{ROG.}  & \textbf{ACC.}         \\ 
			\hline\hline
			SiReRAG-L        & 0.324          & 0.412          & 0.299          & 0.421          & \uline{0.758}  & \uline{0.475}  & 0.647          & \uline{0.673}       & \uline{0.536}  & \uline{0.738}         & 0.260          & 0.426               & \uline{0.308}  & \uline{0.698}         & 0.271          & 0.416               & \uline{0.343}  & \uline{0.661}         \\
			HippoRAG2-L      & \uline{0.433}  & \uline{0.481}  & \uline{0.335}  & \uline{0.425}  & 0.755          & 0.433          & \uline{0.696}  & 0.657               & 0.522          & 0.731                 & \uline{0.280}  & \uline{0.507}       & 0.298          & 0.681                 & \uline{0.288}  & \uline{0.513}       & 0.338          & 0.654                 \\
			\textbf{AGRAG-L} & \textbf{0.441} & \textbf{0.519} & \textbf{0.354} & \textbf{0.436} & \textbf{0.777} & \textbf{0.490} & \textbf{0.745} & \textbf{0.706}      & \textbf{0.545} & \textbf{0.751}        & \textbf{0.293} & \textbf{0.552}      & \textbf{0.312} & \textbf{0.739}        & \textbf{0.297} & \textbf{0.560}      & \textbf{0.351} & \textbf{0.701}        \\ 
			\hline\hline
			% SiReRAG          & 0.324          & \uline{0.328}  & \uline{0.365}  & 0.295          & 0.544          & \uline{0.456}  & \uline{0.754}  & \uline{0.617}       & \uline{0.592}  & \uline{0.560}         & 0.307          & \uline{0.459}       & 0.269          & \uline{0.646}         & \uline{0.311}  & 0.451               & \uline{0.349}  & 0.558                 \\
			% HippoRAG2        & \uline{0.496}  & 0.295          & 0.325          & \uline{0.347}  & \uline{0.563}  & 0.451          & 0.735          & 0.586               & 0.496          & \uline{0.560}         & \uline{0.314}  & 0.453               & \uline{0.332}  & 0.622                 & 0.310          & \uline{0.470}       & 0.343          & \uline{0.598}         \\
			% \textbf{AGRAG}   & \textbf{0.513} & \textbf{0.339} & \textbf{0.386} & \textbf{0.356} & \textbf{0.598} & \textbf{0.589} & \textbf{0.778} & \textbf{0.643}      & \textbf{0.690} & \textbf{0.604}        & \textbf{0.319} & \textbf{0.467}      & \textbf{0.343} & \textbf{0.665}        & \textbf{0.324} & \textbf{0.490}      & \textbf{0.357} & \textbf{0.628}        \\ \hline\hline
		\end{tabular}
	}
	\vspace{-10px}
\end{table*}

\begin{table}
	\centering
	\caption{Low resource experiment efficiency with Qwen3-4B-Instruct-2507 LLM on 2080Ti GPU, where -L denotes low resource setting.}
	\label{tab:exp_low_eff}
	\setlength\tabcolsep{1.5pt}{%
		\begin{tabular}{l|cc|cc|cc|cc} 
			\hline\hline
			\multirow{2}{*}{\textbf{Dataset}} & \multicolumn{4}{c|}{\textbf{Novel}}                                                    & \multicolumn{4}{c}{\textbf{Medical }}                                                                  \\ 
			\cline{2-9}
			& \multicolumn{2}{c|}{\textbf{Time (mins) }} & \multicolumn{2}{c|}{\textbf{AVG.~Token }} & \multicolumn{2}{c|}{\textbf{\textbf{Time (mins)}}} & \multicolumn{2}{c}{\textbf{\textbf{AVG. Token}}}  \\ 
			\hline
			\textbf{Metric}                   & \textbf{Index} & \textbf{Query}            & \textbf{In}    & \textbf{Out}             & \textbf{\textbf{Index}} & \textbf{\textbf{Query}}  & \textbf{\textbf{In}} & \textbf{\textbf{Out}}      \\ 
			\hline\hline
			SiReRAG-L                         & \uline{52.0}   & 183.5                     & \uline{7,898}  & \textbf{1,003}           & 18.1                    & 182.3                    & \textbf{5,113}       & 827                        \\
			HippoRAG2-L                       & 55.7           & \textbf{131.2}            & 7,964          & 1,354                    & \uline{21.8}            & \textbf{162.4}           & 5,577                & \uline{584}                \\
			\textbf{AGRAG-L}                  & \textbf{44.7}  & \uline{132.1}             & \textbf{6,625} & \uline{1,172}            & \textbf{14.4}           & \uline{167.3}            & \uline{5,282}        & \textbf{534}               \\ 
			\hline\hline
			% SiReRAG                           & \uline{85.2}   & 232.6                     & \uline{7,759}  & \textbf{800}             & \uline{40.1}            & 206.8                    & \textbf{4,918}       & 701                        \\
			% HippoRAG2                         & 94.9           & \textbf{134.3}            & 8,138          & \uline{1,073}            & 45.5                    & \textbf{130.4}           & 6,264                & \uline{595}                \\
			% \textbf{AGRAG}                    & \textbf{64.6}  & \uline{142.3}             & \textbf{6,487} & 1,140                    & \textbf{16.2}           & \uline{137.6}            & \uline{5,145}        & \textbf{413}               \\ \hline\hline
		\end{tabular}
	}\vspace{-18px}
\end{table}
To evaluate the efficiency of AGRAG, we compare AGRAG's time and token cost with other baselines' on all 4 dataset tested, the results are shown in Table \ref{tab:efficiency}. 
According to the result, 
AGRAG can also achieve the best time and token efficiency compared with other baselines. 
% It achieves at most 1.66 times faster compared with GraphRAG on Novel dataset, and at least 10.5\% accleration compared to HippoRAG2 on Novel dataset. 
% As for token saving, AGRAG also achieves at most 3.69 times token saving compared with GraphRAG on IFS-REL dataset, and save 9.3\% tokens at least compared with the previous state-of-the-art model HippoRAG2. 

These results lead to two major findings:
% \begin{itemize}[leftmargin=*]
	% \item 
	% , although HippoRAG2 and AGRAG maintain a much larger graph (as shown in Figure \ref{fig:graph_size}) compared to other baselines, they still achieve superior time efficiency. 
	Firstly, both HippoRAG2 and AGRAG achieve superior time efficiency than other baselines, 
	this suggests that frequent LLM calls, particularly those involving long input sequences, are the primary efficiency bottleneck in most Graph-based RAG models. By avoiding the LLM-based graph summarization process, HippoRAG2 and AGRAG significantly alleviate this bottleneck, leading to improved performance.
	
	% \item 
	Secondly, although AGRAG introduces an additional procedure generating the MCMI reasoning graph, the statistics-based entity extraction method avoids the LLM calling.
	Denote the number of nodes in graph as $|\mathcal{V}|$, the number of nodes in text chunk $t$ as $|\mathcal{V}_t|$, the token length of text chunk $t$ as $l_t$, the dimension of LLM hidden state is $d_\text{llm}$, for each test chunk, AGRAG's statistics-based entity extraction achieves $O(|\mathcal{V}|)$ time complexity, and the LLM entity extraction method achieves $O(d_\text{llm}(|\mathcal{V}_t|+l_t)^2)$ complexity. 
	As $d_\text{llm}$ is very large for LLM models (e.g., over 12,000 for GPT-4o \cite{hurst2024gpt}), and $|\mathcal{V}_t|+l_t$ can also be a large number, our AGRAG can still achieve better overall efficiency.
	% The results show that the efficiency gains from this lightweight entity extraction approach—both in terms of token usage and processing time—outweigh the overhead introduced by the MCMI reasoning graph generation. 
% \end{itemize}

	Lastly, according to Table \ref{tab:exp_low} and Table \ref{tab:exp_low_eff}, 
	where we test AGRAG and 2 other best performing baselines with VLLM \cite{kwon2023efficient} deployed Qwen3-4B-Instruct-2507 \cite{yang2025qwen3} LLM on two 2080TI GPUs.  AGRAG-L still can achieve comparable efficiency under low resource setting, while also achieves better accuracy than normal. 
	This is because the Qwen3-4B is a more novel LLM compared with Qwen2.5-14B, it contains richer general knowledge and can perform better in common-sense reasoning \cite{yang2025qwen3}.
	However, due to the model size limitation, Qwen2.5-14B is still better at context understanding and instruction following, making our normal setting outperform the low-resource setting in other metrics.
	% Efficiency wise, while the 14B model was downsized to 4B for the low-resource setting, which lower down the time cost on 2080 Ti GPUs, both settings maintain comparable token costs.

\subsection{Ablation Study}\label{ssec:exp:ablation}
To analyze AGRAG's performance, we conduct an extensive ablation study on GraphRAG-bench tested across 2 datasets and 4 tasks, the ablation models tested are listed as follow:
\begin{itemize}[leftmargin=*]
		\item \textbf{AGRAG w. HR Only}: Here, we apply only AGRAG's hybrid retrieval step, without any reasoning-path generation.
		
	\item \textbf{AGRAG w. LLM ER}: 
	Here, we substitute AGRAG's statistics-based entity extraction method for HippoRAG2's LLM entity extraction approach to test its effects.
	%\footnote{ you claim you are efficient on tf-idf, and llm entity extraction is inefficient. But results show that they do not have too much difference. Your motivation has big problems Yubo: Modified, remove the efficiency claim}

	\item \textbf{AGRAG w.o. HR}: Here, we remove the hybrid retrieved chunks in AGRAG's LLM input, and only input the reasoning subgraph and its involved text chunk to LLM.
	\item \textbf{AGRAG w.o. MCMI}: Here, we simply generate a minimum cost Steiner tree that contains all mapped fact triples as reasoning path, ignoring more complex reasoning structures. 
\item \textbf{AGRAG w.o. Cle.}: Here, we remove the query aware raw triplet filter procedure (Equation \eqref{eq:triplet_filter}) when graph retrieval.
	
% \textbf{AGRAG-L}: The low resource setting of AGRAG, where we apply Qwen-3-4B-Instruct-2507 LLM on 2 2080Ti GPU. 
\end{itemize}

Based on the results in Table \ref{tab:abl_exp}, we can have the following observations regarding AGRAG: 
% \begin{itemize}[leftmargin=*]
%	\item

Firstly, AGRAG w. HR Only achieves the best indexing time efficiency, as it omit all graph related procedures. However, the lack of fine-grained information from graph make it suffer from sub-optimal performance especially on Creative Generation and Contextual Summarization tasks that require summarization to fine-grained information.

	\begin{table}
		\centering
		\caption{Breakdown time cost (mins) of AGRAG on 4 datasets.}
		\label{tab:break_down_efficiency}
		\vspace{-5px}
		\setlength\tabcolsep{1.5pt}{%
			\begin{tabular}{l|c|c|c|c|c|c} 
				\hline\hline
				\textbf{Dataset} & \begin{tabular}[c]{@{}c@{}}\textbf{Entity}\\\textbf{Extraction}\end{tabular} & \begin{tabular}[c]{@{}c@{}}\textbf{Chunk}\\\textbf{Encoding}\end{tabular} & \begin{tabular}[c]{@{}c@{}}\textbf{Triplet}\\\textbf{Filter}\end{tabular} & \begin{tabular}[c]{@{}c@{}}\textbf{MCMI}\\\textbf{Generation}\end{tabular} & \begin{tabular}[c]{@{}c@{}}\textbf{Hybrid}\\\textbf{Retrieval}\end{tabular} & \textbf{QA}  \\ 
				\hline\hline
				Novel            & 62.0                                                                         & 2.8                                                                       & 2.3                                                                      & 3.3                                                                        & 2.4                                                                         & 98.9         \\
				Medical          &                    15.1                                                          &                 0.6                                                          &                           2.8                                                &          9.1                                                                  &                          7.0                                                   &       72.7       \\
				WOS              &           7.4                                                                   &          0.3                                                                 &            3.4                                                               &            1.6                                                                &           0.4                                                                  &     106.7         \\
				IFS-Rel          &        5.6                                                                      &          0.2                                                                 &           1.4                                                                &       0.3                                                                     &        0.1                                                                     &     98.9         \\
				\hline\hline
			\end{tabular}
		}
		\vspace{-10px}
	\end{table}
	 Secondly, although AGRAG w. LLM ER is more time and token-consuming, it still achieves worse performance compared to AGRAG. 
	 % Based on the case study in Section \ref{ssec:exp:case}. 
	 % we find that AGRAG's statistics-based entity extraction method achieves more fine-grained entity recognition compared to the LLM entity extraction approach, generating a larger set of candidate entity nodes for subsequent LLM filtering. As a result, the extracted triples are more natural and semantically complete.
	This is because the LLM entity extraction approach can produce hallucinated outputs that fail to conform to the expected entity extraction format, leading to decoding errors. 
	% In this scenario, current models either skip these entities or text chunks, resulting in incomplete knowledge, or construct graph with the hallucinated outputs, leading to low graph construction quality. 
	% As expected, AGRAG w. LLM ER also exhibits the worst time and token efficiency due to its cumbersome and resource-intensive entity extraction procedure. 
	This highlights the importance of substituting the LLM entity extraction method to avoid hallucination.

	% \item 
	Thirdly, AGRAG w.o. HR experiences a significant performance drop on the Contextual Summarization task and in the coverage score of the Creative Generation task. This is because it provides the LLM with only the reasoning path and its associated text chunks—while efficient, this limits the amount of supportive content available and restricts the model's ability to find sufficient grounded information.

	% \item 
	Fourthly, AGRAG w.o. MCMI performs similarly to AGRAG on the Fact Retrieval and Complex Reasoning tasks but consistently underperforms AGRAG on the more challenging Contextual Summarization and Creative Generation tasks.
	This is because its simple, tree-structured reasoning paths are insufficient to support queries from harder tasks, which may require the inclusion of cyclic structures or multiple answer nodes that are difficult to jointly capture within the Steiner tree framework. 
	
	 Lastly, AGRAG w.o. Cle. achieves suboptimal performance compared to AGRAG, 
	 % According to Table \ref{tab:abl_exp} and Table \ref{tab:break_down_efficiency}, the triplet “cleaning” procedure incurs only a slight increase in time but can improve model effectiveness, 
	 especially in more complex Creative Generation and Contextual Summarization tasks, which require summarization to graph knowledge. 
	 In these scenarios, AGRAG needs the filter procedure to ensure the query-relatedness of retrieved triples, otherwise noise triples are included and affect the model summarization.
	 Considering that omitting the query-related triplet filter can only marginally improve AGRAG's efficiency (reduce less than 1\% of time cost), we consider the triplet cleaning procedure to be worthwhile.
	 % Efficiency wise, while the 14B model was downsized to 4B for the low-resource setting, which lower down the time cost on 2080 Ti GPUs, both settings maintain comparable token costs.

	 We also show the breakdown time cost in Table \ref{tab:break_down_efficiency}, where the main time efficiency bottleneck of AGRAG is the LLM QA procedure, 
	 which were inevitably suffered by all LLM-based RAG methods, 
	 other procedures only lead to a marginal time cost increase. 
	 On Novel dataset, although the entity extraction procedure achieve higher time cost, according to \autoref{tab:abl_exp}, this cost can be even higher if we apply the LLM based entity extraction approach, because AGRAG w. LLM ER (which applies LLM entity extraction method) achieves higher time and token cost than AGRAG.
	 	 
% \end{itemize}
\subsection{Parameter Sensitivity Analysis}
\label{sec:para_sen}

{% \color{blue}
In this subsection, we analyse the parameter sensitivity of AGRAG's two unique hyperparameters: The maximum n-gram $b$ and the entity extraction threshold $\tau$. As all other hyperparameters of AGRAG also exist in other baselines, we just align them with other baselines for fair comparison.
We carry out the analysis on GraphRAG-bench \cite{xiang2025use}, which cover most of the tasks of our experiment.
	
\subsubsection{The maximum n-gram $b$}
%With the data from GraphRAG-bench \cite{xiang2025use}, 
As introduced in Section \ref{sec:gc}, the maximum n-gram $b$ is used as the entity boundary to control the maximum allowed word length of the extracted entities. 
Here, we conduct a test about the maximum n-gram parameter $b\in\{1,3,5,7\}$, as entity distribution in English is, in general, less than 7 words according to \cite{silva2020empirical}.
% According to the experiment in Figure \ref{fig:para_sen}, AGRAG achieves the overall best performance when the entity extraction threshold is set to 0.5 and the maximum n-gram is set to 3 for all tasks. 
From the experiment results in Figure \ref{fig:para_sen}, AGRAG achieves the overall best performance when the maximum n-gram $b$ is set to 3, this is because most of the entities usually consist of no more than 3 words, according to the discovery of \cite{zhu2018gram,fette2007combining,silva2020empirical}. 

For Creative Generation task, AGRAG can achieve better performance with larger maximum n-gram, 
because the correct answer of this task is more flexible than other tasks that have fixed answers or require answers to cover certain facts. 
Therefore, the negative impact of noise introduced by high n-gram phrases is overwhelmed by the additional context they provide, leading to better performance. 
This also highlights AGRAG’s adaptability to different tasks through its flexible graph construction.
In our experiments, we set the maximum n-gram $b$ to 3 for all tasks.

\begin{figure}[t!]
	\centering
	\subfigure[Parameter sensitivity experiment on Novel dataset of GraphRAG-bench.]{
		\label{fig:novel_para}
		\includegraphics[width=\linewidth]{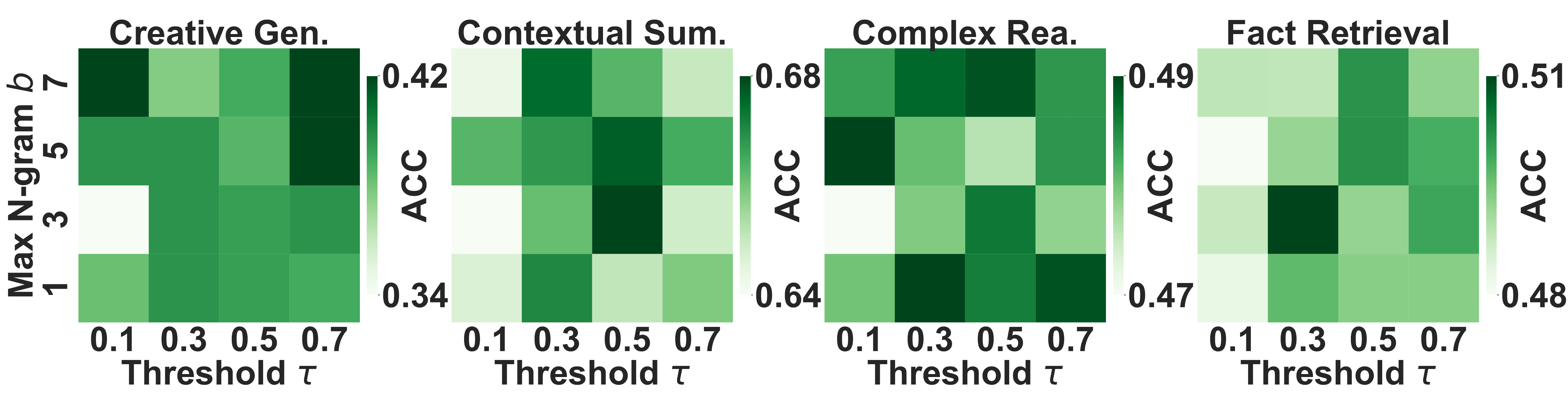}
	}
	\\
	\vspace{-10px}
	% \hspace{-0.3cm}
	\subfigure[Parameter sensitivity experiment on Medical dataset of GraphRAG-bench.]
	{
		\label{fig:medical_para}
		\includegraphics[width=\linewidth]{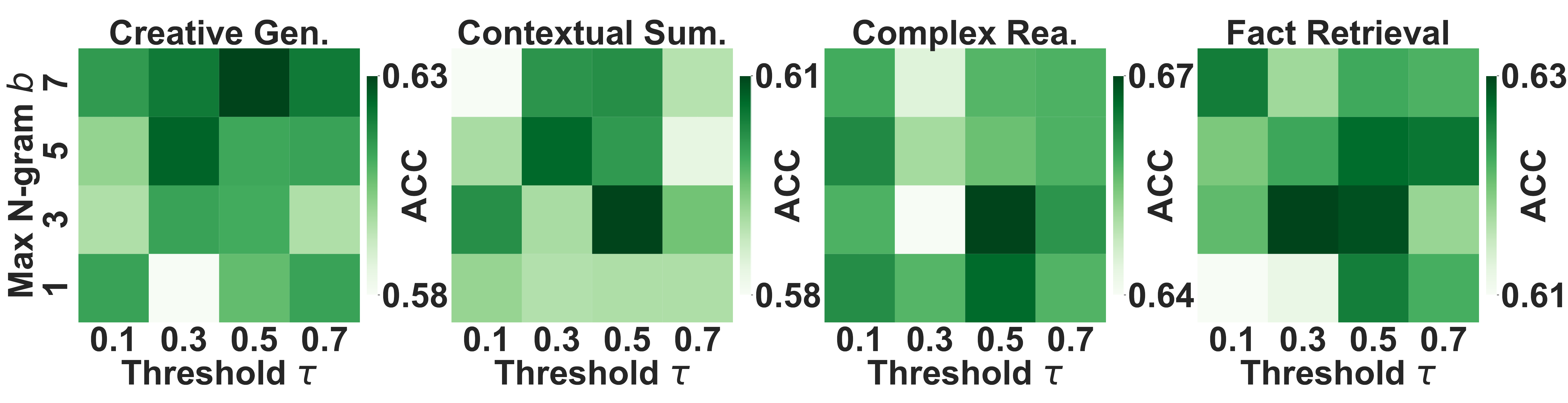}
	}
	\vspace{-5px}
	\caption{Parameter sensitivity experiment on GraphRAG-bench, where x-axis of each heatmap denotes the entity extraction threshold, and y-axis denotes the maximum n-gram. We choose accuracy as the metric; the darker the color of each block, the better the accuracy of the corresponding parameter pair.
	}
	\label{fig:para_sen}
	\vspace{-15px}
\end{figure}
\subsubsection{The entity extraction threshold $\tau$}
As introduced in Section \ref{sec:gc}, in the extraction procedure, we extract words or phrases that have the entity extraction score $\text{ER}(\cdot)>\tau$ as an entity, and we have $0<\text{ER}(\cdot)<1$. 
Considering the computational cost, we only test the entity extraction threshold $\tau\in\{0.1, 0.3, 0.5, 0.7 \}$. 
According to the experiment results in Figure \ref{fig:para_sen}, 
AGRAG typically achieves its best performance when set to 0.3 or 0.5. This is because a large threshold tends to extract less informative entities, and a small one tends to cause insufficient entity extraction; both can affect the model's effectiveness. 
In our experiments, we set the entity extraction threshold $\tau$ to 0.5 for all tasks.

% for better content generation. 
% Such phrases can be better captured by AGRAG’s statistics-based entity extraction under larger n-gram settings, enabling the construction of high-quality nodes in the KG and, consequently, more accurate and faithful reasoning paths in the MCMI subgraph. 
% However, for other tasks, 
% how different tasks impose distinct requirements on the granularity and structure of retrieved information—and demonstrates 
% and MCMI-based reasoning mechanism.

% For the creative generation task on the Novel dataset, as it only contains 67 queries, 
% the For our AGRAG, we set the entity extraction threshold to 0.5 for all tasks,  
% our unique  entity extraction threshold to 0.1, 
% where AGRAG achives the overall best performance according to our parameter sensitivity experiment in Figure \ref{fig:para_sen}.
% F or the entity boundaries, we apply the 3-gram setting, as entities usually consists of no more than 3 words, according to \cite{zhu2018gram,fette2007combining,silva2020empirical}, 
% our parameter sensitivity experiment in Figure \ref{fig:para_sen} can also justify that. 
% It is also worth noting that, on Creative Generation task, a large number of maximum n-gram can benefit the model performance, this is because a large number of maximum n-gram can leads to longer and more entities extracted, hence more comprehensive information reprsented. 
% But on other tasks that require
%\footnote{\#haoyang: write  why  you only have this param experiment: Modified}
%\footnote{\#haoyang: write the reason behind the result Yubo: Modified}
}

\section{Conclusion}\label{sec:conclusion}
% \vspace{-5px}
This paper presents AGRAG, an Advanced Graph-based Retrieval Augmented Generation framework addressing the Inaccurate Graph Construction, Poor Reasoning Ability, and Inadequate Answer issues of current RAG frameworks. 
% AGRAG introduces a statistics-based entity extraction method to replace the LLM entity extraction method applied by current baselines, avoiding hallucination, improving effectiveness, and having better efficiency. 
% By formulating retrieval as a Minimum Cost Maximum Influence (MCMI) subgraph generation problem, AGRAG captures further query-relevant semantic dependencies by allowing more complex graph structure, and the reasoning paths can be more comprehensive and include more query-related graph nodes. 
By providing explicit reasoning paths to help LLM inference, AGRAG makes the LLM better focus on query-related contexts within the retrieved context. 
% These can lead to better performance on complex tasks that require content summarization.
% make the framework able to generalize to complex QA tasks.
% through a three-step process: Graph Indexing, Graph Weighting, and Retrieval \& Generation. 
% We demonstrate the effectiveness and efficiency of KBPearl against the state-ofthe-art techniques, through extensive experiments on realworld datasets.
Experiments on real-world datasets demonstrate the effectiveness and efficiency of AGRAG over state-of-the-art RAG models.
% at most 13.6\% in effectiveness across diverse tasks, and achieves at most 1.66x accleration and 3.69x fewer token usage compared with state-of-the-art Graph-based RAG models in efficiency, significantly reducing computational overhead. Ablation studies confirm that its key components enhance precision, faithfulness, and adaptability to complex scenarios. 
% By integrating structured graph reasoning with LLM generation, AGRAG bridges the gap between knowledge-intensive retrieval and scalable, interpretable RAG systems, laying the groundwork for future advancements in graph modeling and RAG architectures.
\section{Acknowledgements}
Lei Chen is supported by National Key Research and Development Program of China Grant No. 2023YFF0725100, National Science Foundation of China under Grant No. U22B2060, Guangdong-Hong Kong Technology Innovation Joint Funding Scheme Project No. 2024A0505040012, AOE Project AoE/E-603/18, Theme-based project TRS T41-603/20R, CRF Project C2004-21G, Key Areas Special Project of Guangdong Provincial Universities 2024ZDZX1006, Guangdong Province Science and Technology Plan Project 2023A0505030011, HKUST(GZ)  CMCC(Guangzhou Branch) Metaverse Joint Innovation Lab under Grant No. P00659, Hong Kong ITC TC-SKLCRCC26EG01,  ITF grant PRP/004/22FX, Zhujiang scholar program 2021JC02X170, HKUST Webank joint research lab.
Haoyang Li’s work is partially supported by grant No. PolyU P0052504, PolyU-Huawei P0053707, PolyU-Minshang P0057770, PolyU TDG25-28/TBP/11-IICA.
\clearpage
\section{AI-Generated Content Acknowledgement}
In this paper, we use AI models: Qwen3-Max and Gemini3, to polish writing as well as code debugging.
% Traditional 介绍下是什么
% Walking 大写
% E 分成两个subsection，介绍每个parameter是干什么的在哪，范围是什么，结果体现了什么，1,3,5,7 别 写的想reference
% step 2 改为retrieval & graph weighting
\bibliographystyle{IEEEtran}
\bibliography{ICDE2026}
\end{document}